\documentclass[sigconf]{acmart}  % ,authordraft

\usepackage{balance}
\usepackage{algorithm}
\usepackage{algorithmic}
\usepackage{amsfonts}
\usepackage{color}
\usepackage{multirow}

\usepackage{hyperref}
\AtBeginDocument{%
  }

\copyrightyear{2023}
\acmYear{2023}
\setcopyright{acmlicensed}
\acmConference[MM '23]{Proceedings of the 31st ACM International Conference on Multimedia}{October 29-November 3, 2023}{Ottawa, ON, Canada}
\acmBooktitle{Proceedings of the 31st ACM International Conference on Multimedia (MM '23), October 29-November 3, 2023, Ottawa, ON, Canada}
\acmPrice{15.00}
\acmDOI{10.1145/3581783.3612136}
\acmISBN{979-8-4007-0108-5/23/10}

\begin{document}

%%
%% The "title" command has an optional parameter,
%% allowing the author to define a "short title" to be used in page headers.
\title{PNT-Edge: Towards Robust Edge Detection with Noisy Labels by Learning Pixel-level Noise Transitions}

\author{Wenjie~Xuan}
\affiliation{%
    \institution{
    % \fontsize{9.0pt}{0 pt}\selectfont 
    School of Computer Science, National Engineering Research Center for Multimedia Software, Institute of Artificial Intelligence, Hubei Key Laboratory of Multimedia and Network Communication Engineering, Wuhan University}  % , China
    % School of Computer Science, Institute of Artificial Intelligence, Wuhan University}  % , China
    % \institution{Wuhan University}  % 
    \country{}
}
\email{dreamxwj@whu.edu.cn}

\author{Shanshan~Zhao}
\affiliation{% 
    \institution{
    % \fontsize{9.0pt}{2.0 pt}\selectfont 
    JD~Explore~Academy}  % , China
    \country{}
}
\email{sshan.zhao00@gmail.com}

\author{Yu~Yao}
\affiliation{%
    % \fontsize{9.0pt}{0 pt}\selectfont 
    \institution{Mohamed~bin~Zayed~University~of~Artificial~Intelligence \& Carnegie~Mellon~University}  % , Abu Dhabi， America  
    \country{}
}
\email{yu.yao@mbzuai.ac.ae}

\author{Juhua~Liu}
% \authornote{Corresponding Authors: Juhua Liu (email: liujuhua@whu.edu.cn), Bo Du (e-mail: dubo@whu.edu.cn)}
\authornote{Corresponding Authors: Juhua Liu, Bo Du}
\affiliation{%
    % \fontsize{9.0pt}{0 pt}\selectfont 
    \institution{School of Computer Science, National Engineering Research Center for Multimedia Software, Institute of Artificial Intelligence, Hubei Key Laboratory of Multimedia and Network Communication Engineering, Wuhan University}
    % \institution{Wuhan University}
    \country{}
}
\email{liujuhua@whu.edu.cn}

\author{Tongliang~Liu}
\affiliation{%
    % \fontsize{9.0pt}{2.0 pt}\selectfont 
    \institution{The~University~of~Sydney}  % , Australia
    \country{}
}
\email{tongliang.liu@sydney.edu.au}

\author{Yixin~Chen}
\affiliation{%
    % \fontsize{9.0pt}{2.0 pt}\selectfont 
    \institution{Washington~University~in~St.~Louis}  % , America
    \country{}
}
\email{chen@cse.wustl.edu}

\author{Bo~Du}
\authornotemark[1]
\affiliation{%
    % \fontsize{9.0pt}{0 pt}\selectfont 
    \institution{School of Computer Science, National Engineering Research Center for Multimedia Software, Institute of Artificial Intelligence, Hubei Key Laboratory of Multimedia and Network Communication Engineering, Wuhan University}
    % \institution{Wuhan University}
    \country{}
}
\email{dubo@whu.edu.cn}

\author{Dacheng~Tao}
\affiliation{%
    % \fontsize{9.0pt}{2.0 pt}\selectfont 
    \institution{The~University~of~Sydney}  % , Australia
    \country{}
}
\email{dacheng.tao@gmail.com}

%%% official example
% \author{Ben Trovato}
% \authornote{Both authors contributed equally to this research.}
% \email{trovato@corporation.com}
% \orcid{1234-5678-9012}
% \author{G.K.M. Tobin}
% \authornotemark[1]
% \email{webmaster@marysville-ohio.com}
% \affiliation{%
%   \institution{Institute for Clarity in Documentation}
%   \streetaddress{P.O. Box 1212}
%   \city{Dublin}
%   \state{Ohio}
%   \country{USA}
%   \postcode{43017-6221}
% }

%%
%% By default, the full list of authors will be used in the page
%% headers. Often, this list is too long, and will overlap
%% other information printed in the page headers. This command allows
%% the author to define a more concise list
%% of authors' names for this purpose.
% \renewcommand{\shortauthors}{Xuan, }
\renewcommand{\shortauthors}{Wenjie Xuan et al.}

%%
%% The abstract is a short summary of the work to be presented in the
%% article.
\begin{abstract}
    Relying on large-scale training data with pixel-level labels, previous edge detection methods have achieved high performance. However, it is hard to manually label edges accurately, especially for large datasets, and thus the datasets inevitably contain noisy labels. This label-noise issue has been studied extensively for classification, while still remaining under-explored for edge detection. To address the label-noise issue for edge detection, this paper proposes to learn {\bf P}ixel-level {\bf N}oise {\bf T}ransitions to model the label-corruption process. To achieve it, we develop a novel Pixel-wise Shift Learning (PSL) module to estimate the transition from clean to noisy labels as a displacement field. Exploiting the estimated noise transitions, our model, named {\bf PNT-Edge}, is able to fit the prediction to clean labels. In addition, a local edge density regularization term is devised to exploit local structure information for better transition learning. This term encourages learning large shifts for the edges with complex local structures. Experiments on SBD and Cityscapes demonstrate the effectiveness of our method in relieving the impact of label noise. Codes will be available at \href{https://github.com/DREAMXFAR/PNT-Edge}{\textit{github.com/DREAMXFAR/PNT-Edge}}.
    % {\zss This term encourages learning large shifts for the region with complex local structures.} 
    % This term encourages large shifts on latent clean edges with complex local structures via prior knowledge
\end{abstract}

%%
%% The code below is generated by the tool at http://dl.acm.org/ccs.cfm.
%% Please copy and paste the code instead of the example below.
%% updated 0730


%%
%% Keywords. The author(s) should pick words that accurately describe
%% the work being presented. Separate the keywords with commas.
%% A "teaser" image appears between the author and affiliation
%% information and the body of the document, and typically spans the
%% page.

% \begin{teaserfigure}
%   \includegraphics[width=\textwidth]{sampleteaser}
%   \caption{Seattle Mariners at Spring Training, 2010.}
%   \Description{Enjoying the baseball game from the third-base
%   seats. Ichiro Suzuki preparing to bat.}
%   \label{fig:teaser}
% \end{teaserfigure}

% \received{20 February 2007}
% \received[revised]{12 March 2009}
% \received[accepted]{5 June 2009}

%%
%% This command processes the author and affiliation and title
%% information and builds the first part of the formatted document.
\maketitle

%--------------------------------------------------------------------------------------------------------------------------------------------
\section{Introduction}
\begin{figure}[t]
        \centering
        \includegraphics[width=3.1 in]{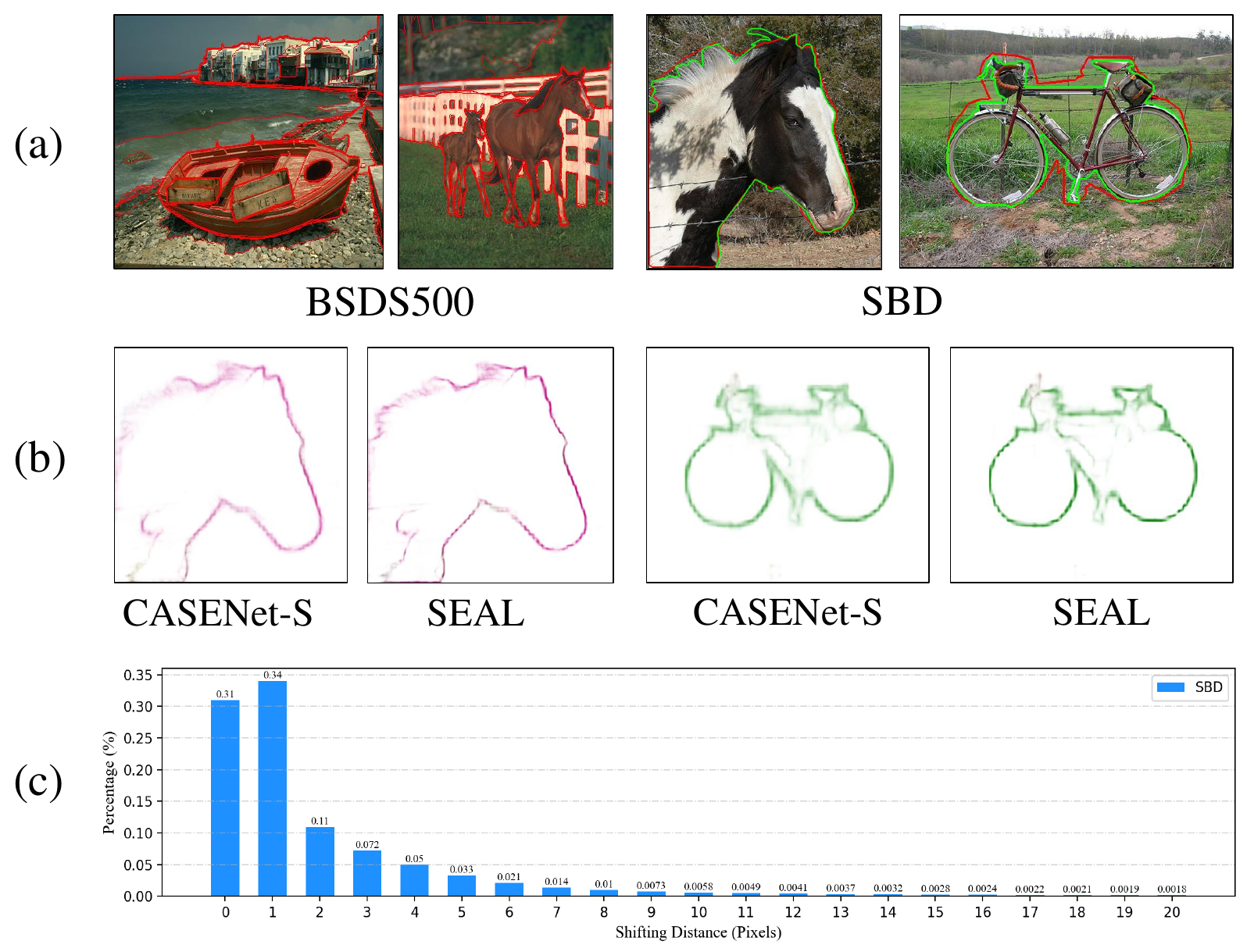}
        \caption{(a) Examples of noisy edge labels. {\color{red}Red} for noisy labels, and {\color{green}green} for clean. BSDS500 only provides noisy labels. (b) Illustration of the impact of label noise. CASENet-S produces more blurred edges than SEAL, especially on detailed structures. (c) Distribution of noisy-label shifts on SBD.}
        \label{fig:examples}
    \end{figure}

     % Background
    Edges can provide useful cues like object shapes and boundaries for high-level vision tasks including semantic segmentation ~\cite{cheng2020boundary}, image generation ~\cite{li2019progressive}, \textit{etc}. Current methods like~\cite{pu2022edter,zhen2020joint, liu2022semantic, xuan2022fcl} have shown remarkable capability in detecting edges, but they rely on large-scale pixel-level annotated training samples. However, obtaining high-quality edge annotations is challenging in real-life scenes. For example, benchmark datasets such as BSDS500~\shortcite{arbelaez2010contour}, SBD~\shortcite{hariharan2011semantic}, and Cityscapes~\shortcite{cordts2016cityscapes} contain different levels of label noise, \textit{i.e.,} misaligned edge annotations, as shown in Fig. \ref{fig:examples}(a).  This issue causes an under-explored challenge in the edge detection community. 

    In Fig. \ref{fig:examples}(c), we calculate pixel-wise shifts between noisy and clean edge labels by minimum-distance matching on an almost clean subset of SBD provided by SEAL~\cite{yu2018simultaneous}. Noisy labels shifting more than 2 pixels are visually perceptive, occupying over 15\%. The imprecise localization of edge pixels, i.e., noisy labels would cause negative impacts on edge detector learning. For example, as shown in Fig. \ref{fig:examples}(b), we can observe CASENet-S trained with noisy labels produces more blurred edges than SEAL with corrected labels. Therefore, it is necessary to develop an effective learning strategy for training robust edge detectors with noisy labels. 

    To our knowledge, few works focus on label noise in edge detection. The seminal attempt is SEAL and following it STEAL~\cite{acuna2019devil} was proposed. To address the label noise, both methods decouple each training step into two stages: 1) edge alignment and label correction, and 2) model training with corrected labels. They achieve edge alignment via solving min-cost graph assignments or level set evolution as an additional optimization step. Better alignment with real edges leads to higher performance. In comparison, this paper considers this label-noise issue from another perspective. For edge detection, label noise mainly results from the trade-off between label quality and efficiency. {\it Therefore, the misaligned noisy labels are not far away from the real edges.} Based on this fact, we wonder whether we can implicitly learn clean labels by modeling label corruption. In fact, learning noise transitions has been proven to be an effective way for label-noise learning in classification~\cite{han2020survey}. Previous works~\cite{liu2015classification,han2020survey} show that when noise transition is given, the model trained on noisy samples converges to the optimal one trained on clean samples with increasing sample size. This paper explores learning noise transitions at pixel level for edge detection.   
    % Previous works like~\cite{liu2015classification} provided theoretical insights to learn classifiers converging to the optimal one trained on clean labels with increasing samples via noise transitions. This paper explores learning noise transitions at the pixel level for edge detection. 

    To arrive at it, we propose {\bf PNT-Edge} towards robust edge detection with noisy labels through learning {\bf P}ixel-level {\bf N}oise {\bf T}ransitions. This transition function describes the label corruption process, \textit{i.e.,} noise transitions. And we achieve it by developing a novel Pixel-wise Shift Learning (PSL) module, which estimates the displacement field of noisy labels via a differentiable STN~\cite{jaderberg2015spatial} structure. Since clean labels are unavailable, it is hard to identify noise transitions merely through noisy labels~\cite{han2020survey}. A common solution is exploiting prior knowledge like ``instances of similar appearance probably have similar transitions" ~\cite{cheng2022instance} to help learn noise transition functions. Since labeling complex edge structures are much harder, such edges tend to contain more label noise as Fig. \ref{hot_map} shows. Considering this fact, we design a local edge density regularization to constrain the structure of the pixel-level noise transitions. Our aim is to encourage large shifts on latent clean edges with complicated local structures. Taking advantage of the designed PSL and the local edge density regularization, our PNT-Edge is able to fit the clean labels implicitly and thus yield thinner and more precise edge maps.
    
    % Experiments results % Contributions
    Experiments prove the effectiveness of PNT-Edge, which improves the ODS-F by 2.4\% and mAP by 4.7\% of the baseline on the re-annotated SBD test set with almost clean labels. We also surpass SEAL ODS-F by 1.3\% and mAP by 1.6\% on SBD. While Cityscapes contains low-level label noise, our method also achieves competitive ODS-F and better mAP compared with existing methods, \textit{i.e.,} 0.1\% higher ODS-F and 4.3\% higher mAP than SEAL. The proposed PNT-Edge can produce visually more precise edge maps than other methods. Our main contributions are summarized as follows:

    \begin{itemize}
        %%% v1
        \item We propose a {\bf PNT-Edge} model to train robust edge detectors with noisy labels by modeling the process of label corruption, \textit{i.e.,} pixel-level noise transition functions. 
        \item We develop a novel PSL module to learn pixel-level noise transitions as a displacement field. And we design a local edge density regularization via prior knowledge to guide the estimation of noise transitions. 
        \item Our PNT-Edge outperforms the SEAL by 1.3\% ODS-F and 1.6\% mAP on SBD. It also obtains competitive ODS-F and 4.3\% higher mAP on Cityscapes than SEAL. Experiments on SBD and Cityscapes validate that our method can relieve the impact of label noise and produce more precise edges. 
    \end{itemize}

    \begin{figure*}[t]
        \centering
        \includegraphics[width=6.2 in]{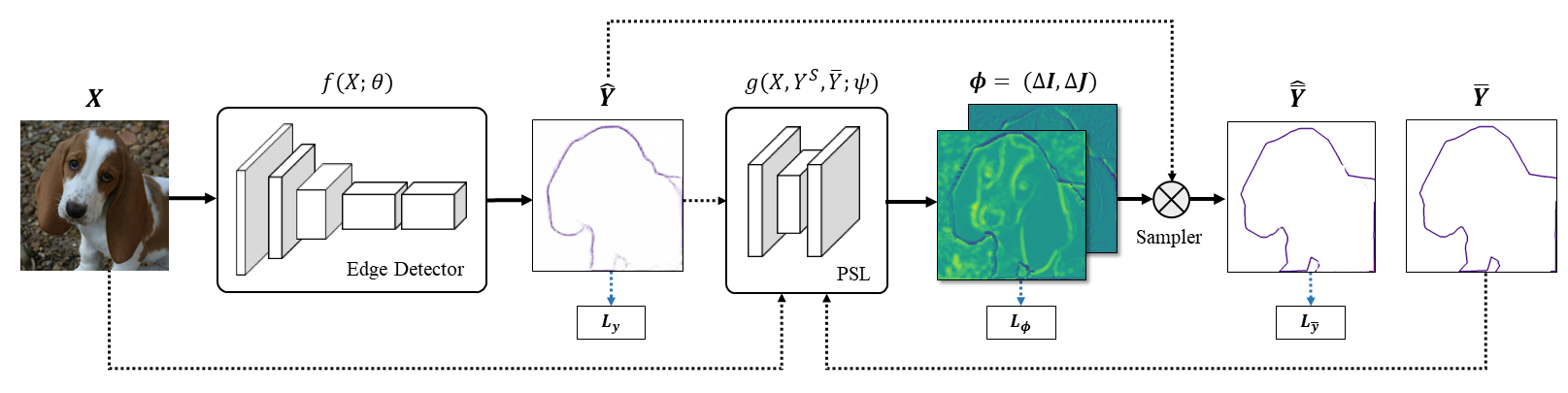}  % 6.4in  \textwidth
        \caption{The architecture of PNT-Edge for edge detection with label noise. It contains an edge detector for edge detection and a PSL module for learning pixel-level noise transitions. We first train edge detector with noisy labels $\bar{Y}$. Then PSL learns $\phi$ for modeling label corruption and adapts prediction $\hat Y$ to $\hat{\bar{Y}}$. Through training edge detector with PSL, $\hat Y$ would finally approximate clean labels. Details refer to Sec. \ref{sec:tedge}. Note that only the edge detector is required for producing the final edge maps.}
        % $L_y, L_\phi, L_{\hat{y}}$ are losses. 
        \label{fig:network}
    \end{figure*}

%--------------------------------------------------------------------------------------------------------------------------------------------
\section{Related Works}
\label{sec: related works}
    \paragraph{Learning with Noisy Labels} Since collecting clean labels for large datasets is expensive, label-noise learning is proposed to train robust models with label noise, especially for classification. Existing methods are generally divided into two categories: (1) Statistically-inconsistent methods employ heuristics like memorization effect~\cite{arpit2017closer} to extract reliable examples~\cite{jiang2018mentornet,malach2017decoupling}, correct noisy labels~\cite{yi2019probabilistic} or design noise-robust loss functions~\cite{patrini2017making}. (2) Statistically-consistent methods are proven to approximate the optimal classifiers trained on clean data~\cite{patrini2017making} if given enough data. But the key challenge for statistically-consistent methods is how to estimate noise transition matrices~\cite{yao2020dual} accurately. Though estimating instance-independent transition matrices has been well-studied under assumptions like anchor points~\cite{liu2015classification}, real-life label noise is instance-dependent and more challenging to identify. Recent works like ~\cite{cheng2022instance,yao2021instance} incorporated prior knowledge and designed regularization to help identify instance-dependent noise transitions.
    % , but such intuitive methods lack theoretical supports
    
    \paragraph{Segmentation with Noisy Labels} Semantic Segmentation also suffers from label noise, especially in semi-supervised learning and medical scenarios. Inspired by the success of label-noise learning in classification, researchers took similar heuristics such as proposing robust loss functions~\cite{wang2020noise} or algorithms for noisy label detection and correction~\cite{shu2019lvc,min2019two}. Specifically, such methods assumed adjacent pixels with similar features share the same label and captured local relationships by dense-CRF~\cite{krahenbuhl2011efficient}, random walk~\cite{bertasius2017convolutional}, {\it etc}. For instance, Yi {\it et al.}~\shortcite{yi2021learning} proposed a graph-based framework to correct noisy labels. Recently, ADELE~\cite{liu2022adaptive} verified the memorization effect for segmentation and proposed a strategy to detect the memorization of each category for label correction. 

    %%% v1
    Edge detection, commonly treated as a segmentation task, is vulnerable to noisy labels because of fine edge structures. Prior methods~\cite{DollarICCV13edges} considered label noise during evaluation by setting a tolerance for matching edges. Later, to deal with multi-annotator bias, RDS~\cite{liu2016learning} proposed to relax labels with Canny~\shortcite{canny1986computational}, and RCF~\cite{liu2017richer} designed an annotator-robust loss. Yang {\it et al.}~\shortcite{Yang_2016_CVPR} refined noisy edges by dense-CRF as pre-processing but it was sensitive to background textures. SEAL~\cite{yu2018simultaneous} first employed a probabilistic model to align noisy labels via min-cost graph assignment and simultaneously trained the detector on refined labels. Similarly, STEAL~\cite{acuna2019devil} utilized a level-set formulation to reason about real edges during training. From a different perspective from SEAL and STEAL, we handle noisy labels by estimating the transitions of noisy labels, so the edge detector can learn real edges implicitly. Thus, we can train robust edge detectors with noisy labels efficiently, without additional individual discrete optimization steps.

%--------------------------------------------------------------------------------------------------------------------------------------------
\section{Methodology}  % Our Approach for Robust Edge Detectors
\label{sec:approach}
    %------------------------------------------------------------------------
    \subsection{Problem Setup}

        % notations for clean dataset
        Following SEAL, this paper focuses on semantic edge detection with $C$ categories, which is commonly treated as a multi-label semantic segmentation task~\cite{yu2017casenet}. Let $(X, Y)\in\mathcal{X}\times\mathcal{Y}$ denote an example drawing from the ideal dataset with clean labels, where the image $X=(x_{ij\cdot})\in R^{H\times W \times 3}$ and the ground-truth $Y= \{ Y_k = (y_{ij}^k) \in R^{H\times W} | k = 1, 2, \cdots, C\}$, $y_{ij}^k\in \{0, 1\}$. Supposing the probabilities of different categories are independent following ~\cite{yu2018simultaneous,acuna2019devil, liu2022semantic}, the goal of semantic edge detection formulates as: 
        % equation
        \begin{equation}
          \max_{\theta}\ P(\hat{Y} | X; \theta) = \max_{\theta}\ \prod_{C}{P(\hat{Y_k} | X; \theta)}, 
          \label{eq:semantic edge}
        \end{equation}
        where $\hat{Y}$ and $\hat{Y}_k$ denote the network predictions, and $\theta$ is the parameter of the semantic edge detector.  

        % considering noisy labels
        However, in real-life scenarios, we can only access a noisy observation $\bar{Y}$ of the ideal clean label $Y$. Therefore, if we train the network with $\bar{Y}$, the detector would eventually fit the noisy posterior. To separate the clean posterior $P(Y_k | X)$, we rewrite Eq. \ref{eq:semantic edge} as conditional probability as follows, 
        \begin{equation}
          \max_{\theta}\ P(\hat{\bar{Y}} | X; \theta) = \max_{\theta_1, \theta_2}\ \prod_{C}{P(\hat{\bar{Y}}_k | \hat{Y}_k, X; \theta_1)} P(\hat{Y}_k | X; \theta_2), 
          \label{eq:noisy edge}
        \end{equation}
        % how to define transition functions
        where $\hat{\bar{Y}}$ and $\hat{\bar{Y}}_k$ denote the prediction of noisy labels, and $\theta_1, \theta_2$ are model parameters. According to Eq. \ref{eq:noisy edge}, if we can estimate the noise transition function $P(\hat{\bar{Y}}_k | \hat{Y}_k, X)$, the clean posterior can be inferred from the noise transition and the noisy posterior. Thus, it is possible to train robust edge detectors implicitly by exploring noise transitions. For label-noise learning in classification, the noise transition matrix between different semantic categories has been widely discussed~\cite{han2020survey}. Here, for studying pixel-level label noise, we define a pixel-level transition function by the displacement field $\phi = (\Delta I, \Delta J) \in R^{H\times W\times 2}$ for learning pixel-wise shifts of noisy labels in edge detection, where $\Delta I = \{\Delta i\}_{H\times W}, \Delta J = \{\Delta j\}_{H\times W}$ represent the pixel-wise shifts of noisy labels along the horizontal and vertical direction, respectively. Here $\phi$ is defined in Euler coordinates~\cite{modersitzki2003numerical}. The relationship between clean and noisy labels formulates as: 
        
        % equation
        \begin{equation}
           Y_k (i,\ j) = \bar{Y}_k (i+ \Delta i, j+ \Delta j).
          \label{eq:shift}
        \end{equation}
        
        % why transition is identifiable v0
        % v1
        Therefore, the key challenge is how to accurately estimate the displacement field $\phi$. Ideally, if both real and noisy labels are accessible, we can identify the displacement field $\phi$ by minimum-distance matching approximately. Because noisy edge labels would not move too far from real edge pixels. However, under the label-noise setting, clean edges are unknown. Fortunately, considering noisy labels only occupy a small part of the dataset, existing works have provided useful methods for extracting confident examples, \textit{i.e.,} probably clean examples, through heuristics like the memorization effect~\cite{arpit2017closer}. Therefore, we first check the memorization effect for edge detectors. Similar to ADELE~\cite{liu2022adaptive}, we select the warm-up model with rapid increments of metrics on the training set and extract confident examples with high confidence~\cite{jiang2018mentornet} (refer to Sec. \ref{sec:memorization}). Then we can approximate the offsets from noisy labels to confident examples by minimum-distance matching. For the other latent clean edges, we can infer their offsets based on the continuity of edges and adjacent confident examples. Moreover, given the fact that complex edges are much harder to label correctly than those with simple structures, we design a local edge density regularization to encourage large shifts on latent clean edges with complicated local structures, which helps constrain the structure of noise transitions. The detailed instructions for the PNT-Edge model are as follows.  
    
    %------------------------------------------------------------------------
    \subsection{PNT-Edge}
    \label{sec:tedge}            
        % overall arch
        As shown in Fig. \ref{fig:network}, our PNT-Edge model consists of two parts: 1) An edge detector $f(\cdot)$ for detecting edges or semantic edges; 2) A PSL module $g(\cdot)$ for estimating pixel-level noise transitions. 

        %------------------------------------------------------------------------
        \paragraph{Edge Detector} While recent works~\cite{zhen2020joint,hu2019dynamic} have set new SOTA for semantic edge detection, we employ CASENet~\cite{yu2017casenet} as our semantic edge detector $f(X; \theta)$ following SEAL for fair comparison. The edge detection process is expressed as: 
        % equation
        \begin{equation}
            \hat{Y} = f(X; \theta). 
            \label{eq:detector}
        \end{equation}

        Note that CASENet can be replaced by other FCN-based edge detectors, and our PNT-Edge is edge-detector-agnostic. 
        % Our method provides a general solution for handling label noise for edge detection. 
        % Specifically, CASENet first treated semantic edge detection as a multi-label task and achieved impressive results.
        
        %------------------------------------------------------------------------
        \paragraph{Pixel-wise Shift Learning Module} To model label corruption according to Eq. \ref{eq:noisy edge}, PSL module takes the image $X$, confident labels $\hat{Y}^S$, and noisy label $\bar{Y}$ to estimate the pixel-level transitions of noisy labels, where $\hat{Y}^S$ is computed by a confidence threshold $\tau$ as: 
        % equation  % \bar{
        \begin{equation}
           \hat{Y}^S = \{\hat{y}^s_{ij}\ |\ p(\hat{y}_{ij}) > \tau, s=1, 2, \cdots, |S|\}.  
          \label{eq:conf}
        \end{equation}
        
        The PSL module $g(X, \hat{Y}^S, \bar{Y}; \psi)$ comprises a localizer and a sampler, where $\psi$ denotes model parameters. The localizer is a 4-layer FCN with shortcuts and outputs displacement field $\phi$. Then the sampler transforms original prediction $\hat{Y}$ of edge detector to $\hat{\bar{Y}}$ according to the field $\phi$ by sampling as:  
        % equation
        \begin{equation}
            \hat{\bar{Y}} = Sampler(\hat{Y}, \phi),\ \phi = g(X, \hat{Y}^S,\bar{Y}; \psi). 
            \label{eq:sampler}
        \end{equation}

        % Since the whole process is differentiable, our PSL module is trainable for learning spatial shifts. 

\begin{figure}[t]
            \centering
                \includegraphics[width=3.2in]{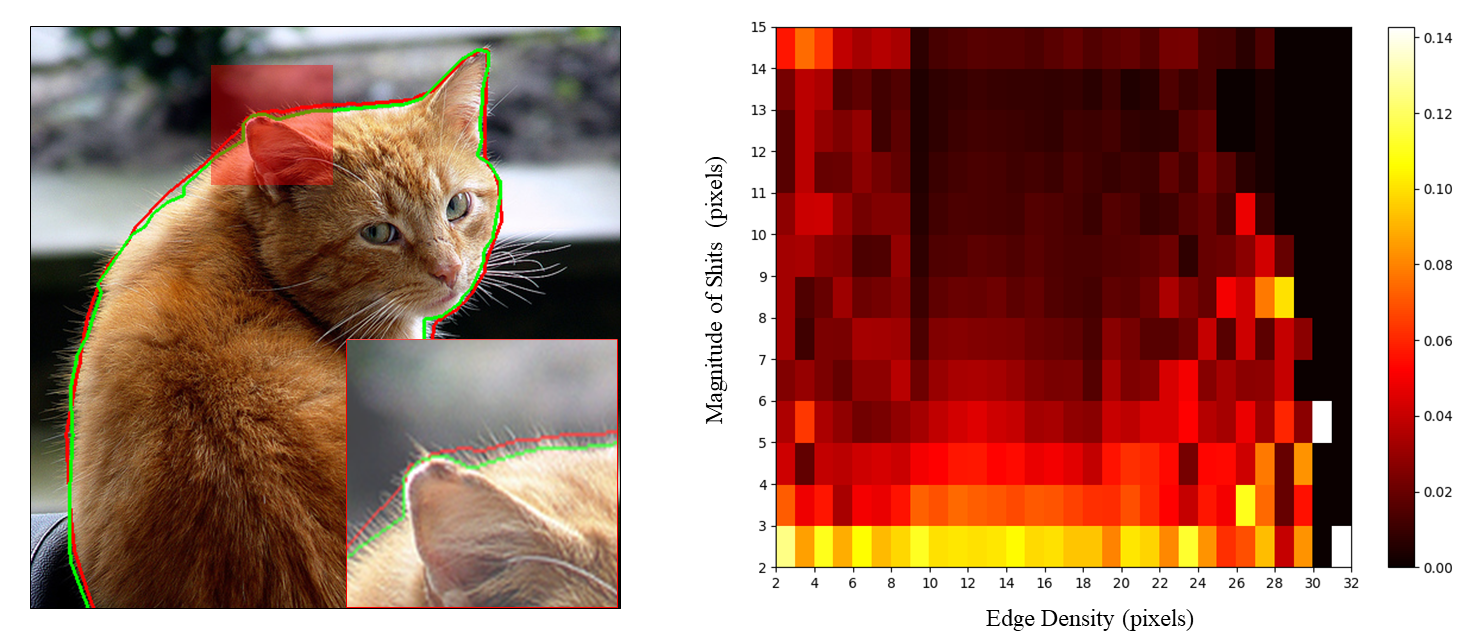}
                \caption{\textcolor{green}{Clean} and \textcolor{red}{noisy} labels, and distribution of noisy-label shifts and edge complexity measured by local edge density. Noisy labels show large offsets on complex structures.}
                % \caption{Distribution of local edge density and shift magnitude. Labels tend to shift more on complex edge structures. }
            \label{hot_map}
        \end{figure}

        %------------------------------------------------------------------------
        \paragraph{Local Edge Density Regularization} Since the edge-label noise is class- and instance-dependent, it is hard to identify the noise transitions without prior knowledge. Generally, complicated edge structures are harder to label correctly because labeling such edges is more costly. This indicates such complex edges are prone to label noise. As a trade-off between label quality and efficiency, noisy labels are mostly over-smoothed. This is a common phenomenon among existing datasets. We verify this fact on SBD by calculating the density of real edges and the magnitude of corresponding shifts. In Fig. \ref{hot_map}, large shifts tend to happen on edges with complicated local structures, \textit{i.e.,} high local edge density, and vice versa. Therefore, to help identify pixel-level noise transitions and guide pixel-wise shift estimation, we utilize this prior knowledge as a structural constraint on the displacement field $\phi$. The proposed local edge density regularization term is defined as: 
        % equation
        \begin{equation}
          L_{dns}(D, C) = \frac{1}{HW} \sum_{i, j}{(d_{ij} - c_{ij})^2}, 
          \label{eq:reg}
        \end{equation}
        % where local edge density $C = \{c_{ij}\}_{H\times W},\ c_{ij} = \frac{\sum{\mathbb{I}(y = 1)}}{N \times N} $ is defined at an $N\times N$ window. Since clean labels are unavailable, we estimate the local edge density with Canny~\cite{canny1986computational}. $D=\{d_{ij}\}_{H\times W},\ d_{ij} = \frac{\sqrt{(\Delta i)^2 + (\Delta j)^2}}{d_{max}}$ is the normalized magnitude measured by Euler distance. Notice that Eq. \ref{eq:reg} does not strictly force $d_{ij} = c_{ij}$, but encourages learning large shifts on latent clean edges with complicated edge structure and vice versa. 
        where local edge density $C = \{c_{ij}\}_{H\times W},\ c_{ij} = \frac{\sum{\mathbb{I}(y = 1)}}{N \times N} $ is defined as the number of edge pixels at an $N\times N$ window centered at $(i,j)$. Since clean labels are unavailable, we estimate the local edge density with Canny~\cite{canny1986computational}. $D=\{d_{ij}\}_{H\times W},\ d_{ij} = \frac{\sqrt{(\Delta i)^2 + (\Delta j)^2}}{d_{max}}$ is the normalized offsets measured by Euler distance, where $d_{max}$ is the maximum offset of each image. Notice that Eq. \ref{eq:reg} does not strictly force $d_{ij} = c_{ij}$, but encourages learning large shifts on latent clean edges with complicated edge structure and vice versa.

        %------------------------------------------------------------------------
        \paragraph{Training Strategy} The whole training pipeline consists of three steps. First, we train the edge detector with noisy labels and select the warm-up model by observing learning curves similar to~\cite{liu2022adaptive}. Second, we extract reliable examples with high confidence and train the PSL module to estimate the displacement field of noisy labels for modeling label corruption. Third, we append the PSL module to the edge detector for joint training. As expressed in Eq. \ref{eq:noisy edge}, since PSL bridges the latent clean labels and noisy labels by the estimated pixel-level transitions, if the noise transitions are estimated accurately, the edge detector $f(X; \theta)$ would eventually fit clean labels. The training pipeline also refers to our Appendix. 

        To train the whole model effectively, carefully designed loss functions are essential. In general, we put constraints on the original edge prediction $\hat{Y}$, displacement field $\phi$ and transformed output $\hat{\bar{Y}}$, abbreviated as $L_y$, $L_{\phi}$ and $L_{\bar{y}}$ in Fig. \ref{fig:network}. In detail, for warm-up training, since SEAL has validated that unweighted loss could bring thinner edge maps and better performance, we employ a multi-label loss without class-balance weight to train CASENet, commonly denoted as CASENet-S, which is computed as: 
        % equation
        \begin{equation}
          L_{edge}(\hat{Y}, \bar{Y}) = \sum_{k}\sum_{i,j} -\bar{y}^k_{ij} log(\hat{y}^k_{ij}) - ( 1-\bar{y}^k_{ij} ) log( 1-\hat{y}^k_{ij} ).
          \label{eq:casenet}
        \end{equation}

        For training the PSL module, we employ three kinds of losses. First, we directly employ the mean squared error to supervise the shifts on confident examples as follows, 
        % equation
        \begin{equation}
            L_{sup}(\phi_S, \phi^{*}_S) = \frac{1}{|S|}\sum_{S}{(\phi (y^s) - \phi^{*}(y^s))^2}, 
            \label{eq:supervise loss}
        \end{equation}
        where the ground-truth shifts $\phi^{*}_S$ are produced by minimum-distance matching between confident labels and noisy labels. Second, since the ideal displacement field would produce $\bar{Y}$ after transformation, we employ a similarity constraint on $\hat{\bar{Y}}$. We empirically find that MSE loss works better than CE in predicting the displacement field, and the similarity loss is formulated as:
        % [original] Second, since the ideal displacement field would produce $\bar{Y}$ after transformation, we employ a similarity loss straightforward as:
        \begin{equation}
            L_{sim}(\hat{\bar{Y}}, \bar{Y}) = \frac{1}{HW}\sum_{i, j}{(\hat{\bar{y}}_{ij} - \bar{y}_{ij})^2}.
            \label{eq:similarity loss}
        \end{equation}
        
        Third, given both clean and noisy edge labels are continuous, the displacement field should be smooth as well. This property helps to infer pixel-wise shifts of edges around confident ones. So, we regularize the field with an L2-loss to encourage smoothness as,  
        \begin{equation}
            L_{smth}(\phi) = || \nabla (\Delta I) ||^2 + || \nabla (\Delta J) ||^2.
            \label{eq:smooth loss}
        \end{equation}

        % % Algorithm
        % \begin{algorithm}[t]
        %     \caption{{\it PNT-Edge} for noisy edge labels learning}
        %     \label{alg:algorithm}
        %     {\bf Input:} Noisy training set $\bar{D}=\{X_i, \bar{Y_i}\}_{i=1}^N$, threshold $\tau$. \ \ \ \ \ \ \ \ \ \ \ \ \ \ \ \\
        %     {\bf Output:} Noise-robust edge detector $f(X; \hat{\theta})$.\ \ \ \ \ \ \ \ \ \ \ \ \ \ \ \ \ \ \ \ \ \ \ \ \ \ \ \ \ \
            
        %     \begin{algorithmic}[1] %[1] enables line numbers
        %         \STATE $//$ {\it Step 1: Warm-up training}
        %         \STATE Train the edge detector on the noisy dataset $\bar{D}$ to obtain the initial detector $f(X; \theta_0)$ with Eq. \ref{eq:casenet}.
        %         \STATE $//$ {\it Step 2: Train the PSL module}
        %         \FOR{$i=1$ to $N$}
        %             \STATE Extract confident labels $Y^S_i$ as Eq. \ref{eq:conf} with $f(X; \theta_0)$; 
        %             \STATE Generate field $\phi$ and transformed output $\hat{\bar{Y}}$ as Eq. \ref{eq:sampler}; 
        %             \STATE Compute the loss as Eq. \ref{eq:psl loss};
        %             \STATE Optimize the parameter $\psi$ by SGD; 
        %         \ENDFOR
        %         \STATE \textbf{return} $\hat{\psi}$
                
        %         \STATE $//$ {\it Step 3: Train the edge detector with PSL}
        %         \FOR{$n=1$ to $N$}
        %             \STATE Generate predictions $\hat{Y}_i, \hat{\bar{Y}}_i$ by $f\circ g$ as Eq. \ref{eq:detector} and Eq. \ref{eq:sampler}; 
        %             \STATE Compute the loss as Eq. \ref{eq:joint loss};
        %             \STATE Optimize the parameter $\theta$ by SGD; 
        %         \ENDFOR
                
        %         \STATE \textbf{return} $\hat{\theta}$
        %     \end{algorithmic}
        % \end{algorithm}
        
        To help identify the noise transition function, our local edge density regularization is added and implemented as Eq. \ref{eq:reg}. When conducting experiments, we find the proposed $L_{dns}$ also regularizes the smoothness of the displacement field. Details refer to Sec. \ref{sec:ablation}. Therefore, the overall loss for PSL training is formulated as follows, where $\{ \alpha_i \}_{i=1}^4$ are loss weights.  
        \begin{equation}
            L_{PSL} = \alpha_{1} L_{sup} + \alpha_{2} L_{sim} + \alpha_{3} L_{smth} + \alpha_{4} L_{dns}.
            \label{eq:psl loss}
        \end{equation}

        For the joint training of edge detector and PSL, we first intuitively supervise the final output $\hat{\bar{Y}}$ by the noisy label $\bar{Y}$ as follows, 
        \begin{equation}
            L_{edge}(\hat{\bar{Y}}, \bar{Y}) = \sum_{k}\sum_{i,j} -\bar{y}^k_{ij} log(\hat{\bar{y}}^k_{ij}) - ( 1-\bar{y}^k_{ij} ) log( 1-\hat{\bar{y}}^k_{ij} ).
            \label{eq:joint edge}
        \end{equation}

        \begin{figure}[t]
            \centering
                \includegraphics[width=3.0 in]{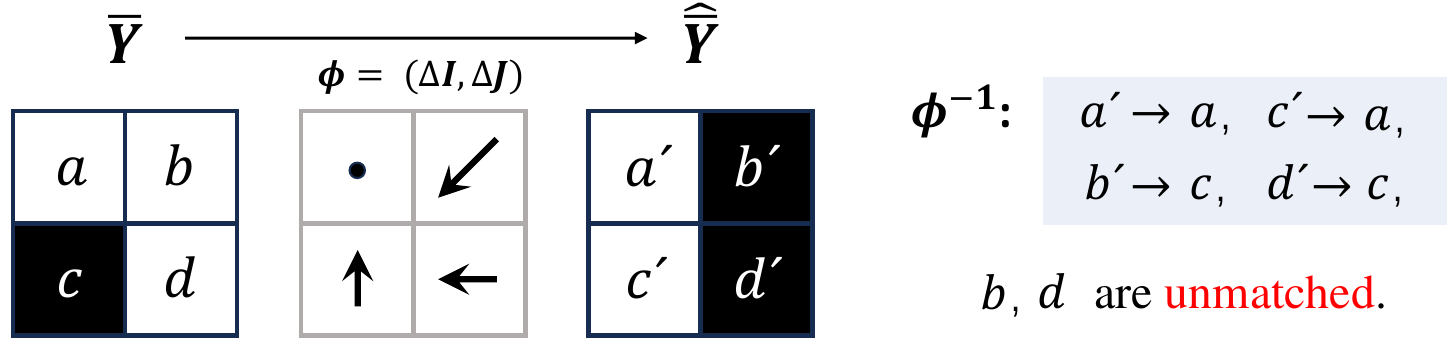}
                \caption{Illustration of unmatched pixels and $\phi^{-1}$ in Eq.14.}  
                % The unmatched pixels are non-edge pixels, where the predictions should be zero. 
            \label{fig:eq14}
        \end{figure}
        % v1 Second, we force the predictions of unmatched pixels to be zero, which is computed as:   
        Additionally, we force the predictions of unmatched pixels to be zero. Because $\phi$ is defined as a many-to-one mapping, there exist unmatched pixels in the output $\hat{Y}$ as illustrated in Fig.\ref{fig:eq14}. If only the final output $\hat{\bar{Y}}$ is supervised, the predictions of such unmatched pixels will be uncontrolled, which is undesirable. Experimental evidence refers to Sec. \ref{sec:ablation}. The unmatched loss is computed as:   
        \begin{equation}
            L_{um} = \sum_k \sum_{(i,j) \notin \phi^{-1}} -log( 1 - \hat{y}^k_{ij}), 
            \label{eq:joint match}
        \end{equation}   
        where $\phi^{-1}$ denotes the generalized inverse of $\phi$. Thus, the overall loss for joint training is as follows:
        \begin{equation}
            L_{joint} = \beta_{1} L_{edge} + \beta_{2} L_{um}, 
            \label{eq:joint loss}
        \end{equation}
        where $\{ \beta_i \}_{i=1}^2$ are weights. Since both losses are equally important, we empirically set $\beta_1 = \beta_2 = 1$ for simplicity.        

    %------------------------------------------------------------------------
    \subsection{Discussion}
        %%% v0 
        In this part, we further analyze our work by making comparisons with some previous related works.
               
        First, our method differs from label-noise learning for classification~\cite{han2020survey}, where statistically-consistent methods~\cite{patrini2017making} proposes to estimate the noise transitions between categories. In comparison, aiming at edge detection with label noise, this paper explores noise transition functions at pixel level through the displacement field. We also incorporate prior knowledge of edge-label noise and design a regularization term for learning noise transitions better.  

        Second, it is worth noting that our method varies from current works on addressing label noise in segmentation. These methods attempt to alleviate the issue through early stopping~\cite{liu2022adaptive}, prediction consistency~\cite{min2019two}, modeling local affinity by GAT~\cite{yi2021learning}, low-level cues~\cite{shu2019lvc}, {\it etc}. Such methods rely on the affinity between boundary and interior pixels for label correction, but it is unavailable in edge detection. Moreover, noise transition is rarely explored in segmentation. This paper aims to implicitly learn clean labels by modeling noisy label corruption, rather than label refinements. 
        % via noise transitions 
        % there are current works addressing label noise in segmentation. 
            
        Third, learning pixel offsets is often exploited to refine predictions. For example, SharpContour~\cite{zhu2022sharpcontour} proposed a contour-based model to improve boundary quality for instance segmentation as post-processing. Recent E2EC~\cite{zhang2022e2ec} proposed a global contour deformation module and a multi-direction alignment scheme to optimize boundaries for contour-based instance segmentation. DELSE~\cite{wang2019object} predicted pixel motions and deformed the contour by curve evolution for interactive object segmentation. Different from these efforts, we attempt to implicitly learn robust edge detectors by employing pixel offsets as transitions between clean and noisy labels.            

%--------------------------------------------------------------------------------------------------------------------------------------------
\section{Experiments}
\label{sec:experiments}
    
    %------------------------------------------------------------------------   
    \subsection{Datasets}
        % sbd
        \paragraph{Semantic Boundary Dataset (SBD)} It contains 11,355 images from the PASCAL VOC 2011 trainval set with category- and instance-level semantic edges of 20 classes. There are 8,498 images for training and 2,857 images for testing. SEAL provided a re-annotated sub-test set of 1,059 images with almost clean labels for evaluation. We randomly sample 1,000 images from the training set and 50\% of the sub-test set for validation. Here we only focus on instance-sensitive edges following SEAL and STEAL. 
        %  following the PASCAL VOC definition

        % cityscapes
        \paragraph{Cityscapes} It contains 2,975 road images for training, 500 for validation, and 1,525 for testing. SEAL provided edge labels. We test on the validation set following SEAL. Since Cityscapes is annotated by experts, it contains much less label noise than SBD. Considering that Cityscapes does not provide clean labels, we evaluate performance on noisy labels with relaxed criteria following SEAL.  
        
    %------------------------------------------------------------------------    
    \subsection{Evaluation Metrics}
        We employ the same metrics as SEAL and test on clean labels. We report maximum F-measure at optimal dataset scale (ODS-F) and mAP, and set the matching tolerance to 0.0075 ($\approx$4 pixels) for SBD and 0.0035 ($\approx$8 pixels) for Cityscapes. Following SEAL, we employ two settings: 1) ``Thin" for matching the thinned prediction with ground truth, and 2) ``Raw" for matching the raw prediction with ground truth. Our method does not employ NMS post-processing. 
        % , \textit{i.e.} a multi-label extension of the BSDS benchmark
        % for post-processing
    
    \begin{figure}[b]
        \centering
        \includegraphics[width=3.2 in]{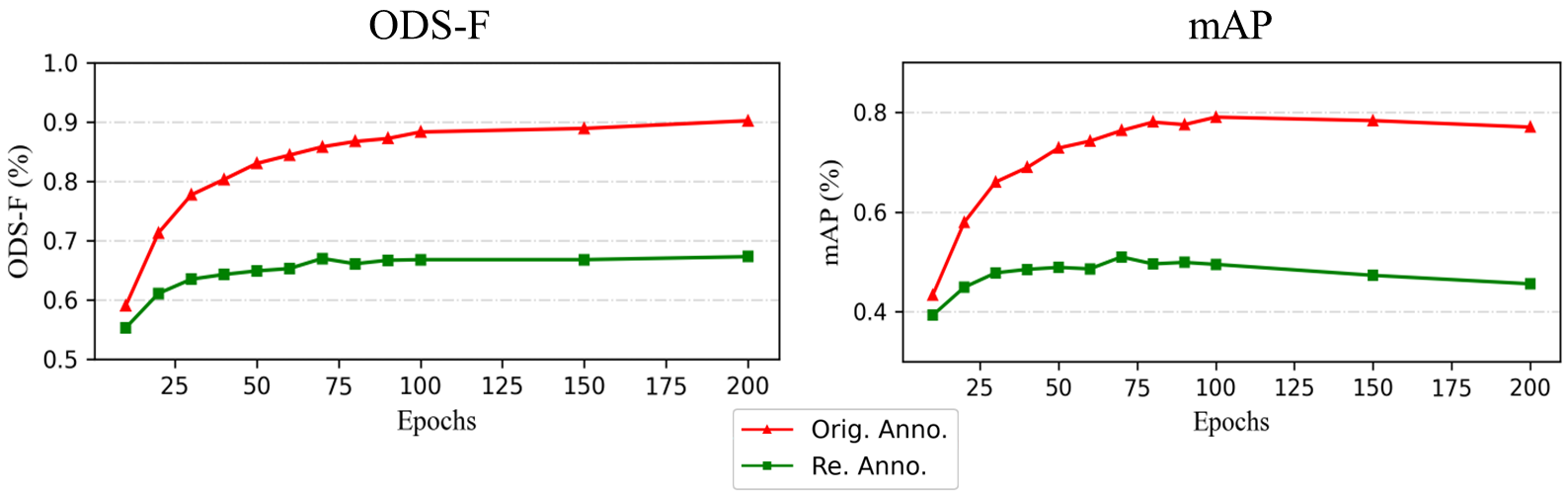}
        \caption{Evaluation protocols on the sub-test of SBD that validate the memorization effect in edge detection.}
        \label{fig:memory}
    \end{figure}
    
    %------------------------------------------------------------------------  
    \makeatletter
        \newcommand{\ssymbol}[1]{^{\@fnsymbol{#1}}}
    \makeatother
    % SBD performance table 
    \begin{table*}[t]
        \centering
        \renewcommand{\arraystretch}{1.0}
        \resizebox{\textwidth}{!}{
        \begin{tabular}{c|c|c|cccccccccccccccccccc|c}
            \toprule
            Setting    & Metric   & Method   & aero     & bike     & bird     & boat     & bottle     & bus     & car     & cat     & chair     & cow     & table     & dog     & horse     & mbike     & person     & plant     & sheep     & sofa     & train     & tv     & AVG.     \\
            \midrule
            % CASENet
            \multirow{10}{*}{Thin}    & \multirow{5}{*}{ODS-F}   & CASENet   & 0.743     & 0.597     & 0.732     & 0.478     & 0.668     & 0.787     & 0.673     & 0.760     & 0.475     & 0.696     & 0.361     & 0.756     & 0.726     & 0.613     & 0.746     & 0.425     & 0.715     & 0.487     & 0.716     & 0.552     & 0.635     \\
            % CASENet-S
            ~    & ~   & CASENet-S   & 0.759     & 0.663     & 0.755     & 0.519     & 0.664     & 0.798     & 0.710     & 0.788     & 0.501     & 0.698     & 0.397     & 0.772     & 0.746     & 0.649     & 0.769     & 0.471     & 0.727     & 0.515     & 0.727     & 0.573     & 0.658     \\
            % SEAL
            ~    & ~   & SEAL   & 0.781     & 0.659     & 0.767     & 0.524     & 0.684     & 0.800     & 0.706     & \bf 0.796     & 0.500     & 0.726     & 0.414     & 0.782     & 0.750     & 0.655     & \bf 0.784     & 0.492     & 0.730     & \bf 0.522     & 0.739     & 0.582     & 0.669     \\
            % STEAL
            ~    & ~   & STEAL$\ssymbol{2}$   & \bf 0.790     & 0.658     & \bf 0.773     & 0.546     & 0.686     & \bf 0.815     & 0.711     & 0.784     & \bf 0.523     & 0.737     & 0.428     & \bf 0.792     & \bf 0.764     & \bf 0.668     & 0.782     & 0.491     & 0.752     & 0.500     & \bf 0.749     & 0.594     & 0.677     \\
            \cline{3-24}
            % CASENet-reproduction
            % ~    & ~   & CASENet-S$^*$   & 0.757     & 0.664     & 0.772     & 0.502     & \bf 0.698     & 0.814     & \bf 0.719     & 0.784     & 0.496     & \bf 0.750     & 0.375     & 0.769     & 0.732     & 0.655     & 0.781     & 0.481     & 0.768     & 0.503     & 0.746     & \bf 0.619     & 0.669     \\
            % Ours
            % ~    & ~   & Ours(tmp)   & 0.787     & \bf 0.679     & 0.768     & \bf 0.549     & 0.692     & 0.807     & 0.712     & 0.783     & 0.506     & \bf 0.751     & \bf 0.431     & 0.778     & 0.756     & 0.664     & 0.780     & \bf 0.519     & \bf 0.782     & 0.511     & 0.745     & 0.607     & \bf 0.680     \\
            ~    & ~   & \bf Ours   & 0.787     & \bf 0.680     & 0.770     & \bf 0.547     & \bf 0.697     & 0.813     & \bf 0.714     & 0.786     & 0.503     & \bf 0.750     & \bf 0.436     & 0.779     & 0.757     & 0.666     & \bf 0.784     & \bf 0.518     & \bf 0.784     & 0.516     & 0.748     & \bf 0.608     & \bf 0.682     \\
            \cline{2-24}
            % CASENet-MAP
            ~    & \multirow{5}{*}{mAP}   & CASENet   & 0.537     & 0.444     & 0.479     & 0.309     & 0.489     & 0.596     & 0.523     & 0.629     & 0.360     & 0.494     & 0.253     & 0.591     & 0.498     & 0.423     & 0.599     & 0.272     & 0.531     & 0.409     & 0.490     & 0.374     & 0.465     \\
            % CASENet-S-mAP
            ~    & ~   & CASENet-S   & 0.677     & 0.519     & 0.690     & 0.406     & 0.627     & 0.735     & 0.636     & 0.753     & 0.409     & 0.606     & 0.304     & 0.722     & 0.654     & 0.558     & 0.730     & 0.332     & 0.630     & 0.449     & 0.671     & 0.484     & 0.580     \\
            % SEAL-mAP
            ~    & ~   & SEAL   & 0.733     & 0.568     & 0.720     & 0.426     & 0.661     & 0.754     & 0.661     & \bf 0.781     & 0.420     & 0.659     & \bf 0.334     & 0.757     & 0.689     & 0.575     & 0.764     & 0.376     & 0.675     & \bf 0.468     & 0.692     & 0.513     & 0.611     \\
            % STEAL-mAP
            ~    & ~   & STEAL$\ssymbol{2}$   & 0.747     & 0.595     & \bf 0.742     & 0.436     & 0.658     & \bf 0.775     & \bf 0.677     & 0.774     & \bf 0.428     & \bf 0.707     & 0.315     & \bf 0.775     & \bf 0.750     & \bf 0.607     & 0.773     & 0.381     & 0.701     & 0.389     & \bf 0.714     & 0.503     & 0.622     \\
            % CASENet-S reproduction -mAP
            \cline{3-24}
            %  ~    & ~   & CASENet-S$^*$   & 0.667     & 0.555     & 0.697     & 0.390     & 0.660     & 0.753     & 0.653     & 0.742     & 0.397     & 0.676     & 0.250     & 0.716     & 0.658     & 0.547     & 0.743     & 0.365     & 0.669     & 0.432     & 0.671     & \bf 0.545     & 0.589     \\
            % Ours-mAP
            % ~    & ~   & Ours   & \bf 0.763     & \bf 0.639     & 0.740     & \bf 0.449     & \bf 0.672     & \bf 0.764     & 0.676     & 0.774     & 0.414     & 0.706     & 0.332     & 0.761     & 0.734     & \bf 0.615     & \bf 0.783     & \bf 0.442     & \bf 0.718     & 0.428     & 0.704     & \bf 0.545     & \bf 0.633     \\
            ~    & ~   & \bf Ours   & \bf 0.750     & \bf 0.619     & 0.728     & \bf 0.440     & \bf 0.673     & 0.761     & 0.665     & 0.771     & 0.412     & 0.699     & \bf 0.334     & 0.753     & 0.725     & 0.606     & \bf 0.781     & \bf 0.429     & \bf 0.714     & 0.434     & 0.699     & \bf 0.541     & \bf 0.627     \\
            \midrule % -------------------------------------------------------------------------------------
            \multirow{10}{*}{Raw}    & \multirow{5}{*}{ODS-F}   & CASENet   & 0.657     & 0.514     & 0.649     & 0.429     & 0.572     & 0.682     & 0.582     & 0.659     & 0.453     & 0.597     & 0.329     & 0.640     & 0.657     & 0.524     & 0.654     & 0.408     & 0.650     & 0.428     & 0.613     & 0.477     & 0.559     \\
            % CASENet-S-raw-F
            ~    & ~   & CASENet-S   & 0.689     & 0.557     & 0.709     & 0.473     & 0.619     & 0.715     & 0.647     & 0.712     & 0.480     & 0.648     & 0.372     & 0.691     & 0.688     & 0.581     & 0.702     & 0.442     & 0.687     & 0.461     & 0.657     & 0.525     & 0.603     \\
            % SEAL-raw-F
            ~    & ~   & SEAL   & 0.753     & 0.605     & 0.752     & 0.512     & 0.654     & 0.761     & 0.679     & 0.760     & 0.497     & 0.694     & 0.399     & 0.747     & 0.728     & 0.621     & 0.741     & 0.482     & 0.723     & 0.492     & 0.705     & 0.566     & 0.644     \\
            % STEAL-raw-F
            ~    & ~   & STEAL$\ssymbol{2}$   & 0.709     & 0.559     & 0.716     & 0.476     & 0.616     & 0.726     & 0.646     & 0.702     & 0.475     & 0.674     & 0.373     & 0.706     & 0.694     & 0.591     & 0.692     & 0.443     & 0.691     & 0.426     & 0.677     & 0.535     & 0.606     \\
            \cline{3-24}
            % casenet-s repro
            % ~    & ~   & CASENet-S$^*$   & 0.704     & 0.592     & 0.723     & 0.469     & 0.643     & 0.738     & 0.662     & 0.724     & 0.469     & 0.722     & 0.339     & 0.702     & 0.661     & 0.567     & 0.728     & 0.460     & 0.725     & 0.441     & 0.668     & 0.582     & 0.616     \\
            % Ours
            % ~    & ~   & Ours(tmp)   & 0.772     & 0.671     & 0.769     & 0.551     & 0.675     & 0.776     & 0.700     & 0.763     & 0.512     & 0.735     & 0.423     & 0.760     & 0.756     & 0.655     & 0.774     & 0.525     & 0.781     & 0.494     & 0.725     & 0.601     & 0.671     \\
            ~    & ~   & \bf Ours   & \bf 0.766     & \bf 0.660     & \bf 0.767     & \bf 0.543     & \bf 0.673     & \bf 0.776     & \bf 0.696     & \bf 0.766     & \bf 0.505     & \bf 0.734     & \bf 0.423     & \bf 0.758     & \bf 0.751     & \bf 0.644     & \bf 0.771     & \bf 0.517     & \bf 0.781     & \bf 0.493     & \bf 0.717     & \bf 0.598     & \bf 0.667     \\
            \cline{2-24}
            % CASENet mAP raw
            ~    & \multirow{5}{*}{mAP}   & CASENet   & 0.668     & 0.509     & 0.596     & 0.346     & 0.510     & 0.675     & 0.563     & 0.680     & 0.416     & 0.547     & 0.263     & 0.657     & 0.668     & 0.480     & 0.701     & 0.348     & 0.618     & 0.390     & 0.615     & 0.423     & 0.534     \\
            % CASENet-S-RAW-Map
            ~    & ~   & CASENet-S   & 0.755     & 0.575     & 0.750     & 0.455     & 0.625     & 0.755     & 0.662     & 0.762     & 0.456     & 0.657     & 0.309     & 0.738     & 0.730     & 0.598     & 0.763     & 0.391     & 0.701     & 0.420     & 0.700     & 0.504     & 0.615     \\
            % seal-raw-mAP
            ~    & ~   & SEAL   & \bf 0.810     & 0.634     & \bf 0.792     & \bf 0.496     & \bf 0.668     & \bf 0.794     & \bf 0.699     & \bf 0.810     & \bf 0.472     & 0.705     & \bf 0.344     & \bf 0.797     & \bf 0.766     & 0.633     & \bf 0.803     & 0.449     & 0.736     & \bf 0.452     & \bf 0.739     & \bf 0.539     & \bf 0.657     \\
            % STEAL-raw-mAP
            ~    & ~   & STEAL$\ssymbol{2}$   & 0.756     & 0.559     & 0.749     & 0.432     & 0.619     & 0.751     & 0.653     & 0.746     & 0.419     & 0.677     & 0.287     & 0.744     & 0.736     & 0.592     & 0.740     & 0.382     & 0.695     & 0.354     & 0.710     & 0.493     & 0.605     \\
            \cline{3-24}
            % casenet-s RE -raw-mAP
            % ~    & ~   & CASENet-S$^*$   & 0.743     & 0.610     & 0.748     & 0.433     & 0.651     & 0.768     & 0.673     & 0.772     & 0.437     & \bf 0.736     & 0.238     & 0.739     & 0.702     & 0.574     & 0.781     & 0.429     & 0.730     & 0.403     & 0.701     & \bf 0.561     & 0.622     \\
            % line
            % ~    & ~   & Ours(tmp)   & 0.788     & 0.668     & 0.775     & 0.496     & 0.662     & 0.777     & 0.692     & 0.794     & 0.435     & 0.724     & 0.333     & 0.774     & 0.762     & 0.640     & 0.799     & 0.483     & 0.767     & 0.410     & 0.740     & 0.557     & 0.654     \\
            ~    & ~   & \bf Ours   & 0.790     & \bf 0.662     & 0.774     & 0.492     & 0.664     & 0.784     & 0.688     & 0.797     & 0.434     & \bf 0.727     & 0.335     & 0.775     & 0.762     & \bf 0.634     & 0.801     & \bf 0.477     & \bf 0.769     & 0.414     & 0.736     & 0.555     & 0.653     \\
            \bottomrule
        \end{tabular}
        }
        \caption{Comparisons with previous SOTA methods on the re-annotated SBD test set. ``AVG.'' and $\ssymbol{2}$ denote category-averaged performance and using NMS for post-processing, respectively. }  % $*$ denotes our reproduced results. 
        \label{tab:sbd}
    \end{table*}

    % visualization
    \begin{figure*}[h]
        \centering
        \includegraphics[width=6.0 in]{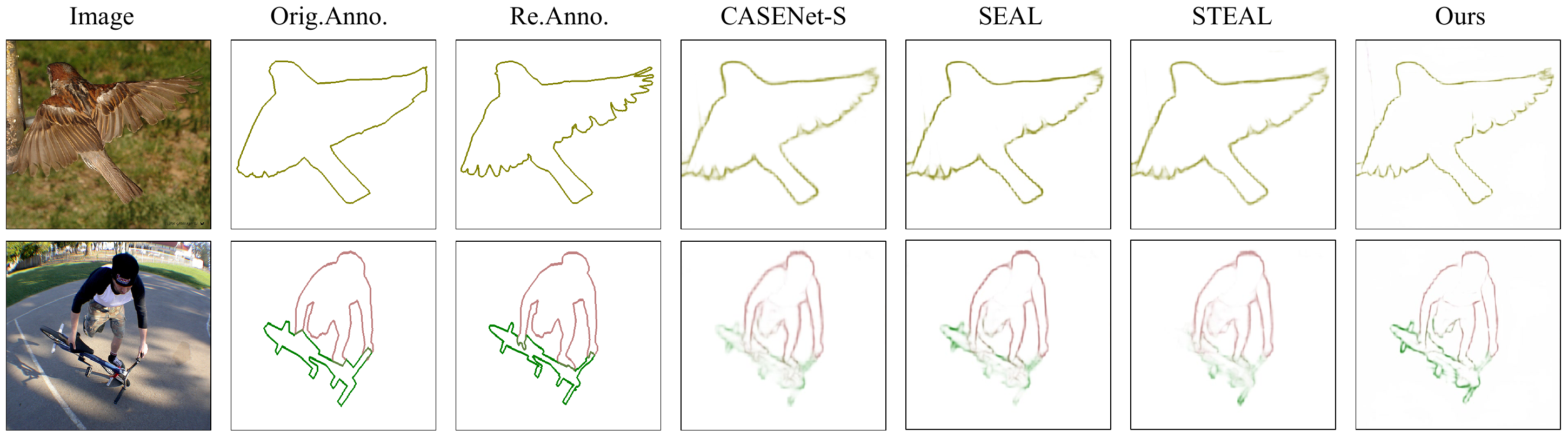}
        \caption{Example results on SBD. Our method generates more precise edge maps with clear details.}
        \label{fig:sbd_vis}
    \end{figure*}

    %------------------------------------------------------------------------    
    \subsection{Implementation Details}
    \label{subsec: implementation}
        %We use PyTorch for implementation. 
        Following SEAL, we crop images by $472\times 472$ for training and employ scaling and random flipping. We initialize CASENet with pre-trained parameters provided by SEAL and randomly initialize PSL module. For warm-up training, we employ a learning rate 5e-8$/$2.5e-8 and train $30$ epochs for SBD$/$Cityscapes with batch size 8 and learning rate decay. For training PSL, we set the learning rate as 1e-6 and freeze CASENet. We train PSL module for 10 epochs and decay the learning rate by $0.1$ every $10k$. For joint training, we initialize CASENet with warm-up parameters and freeze PSL module. Then we change the learning rate to 1e-9 and train for another 10 epochs while keeping other settings the same. For hyper-parameters, we empirically set $\tau=0.1, \alpha_1=0.01, \alpha_2=1, \alpha_3=0, \alpha_4=3$ according to Sec. \ref{sec:ablation}. All experiments are conducted on NVIDIA GeForce RTX 3090 using PyTorch.

    %------------------------------------------------------------------------    
    \subsection{Memorization Effect in Edge Detection}
    \label{sec:memorization}
        %%% v1
        We first revisit memorization effect in edge detection. Since clean training samples are required for evaluation, we divide the re-annotated SBD test set into 847$/$212 images for train/test and train CASENet-S with noisy labels. Fig. \ref{fig:memory} reports the performance on noisy and clean labels. While metrics on noisy labels keep increasing, metrics on clean labels only increase at early stages. It verifies detectors first fit clean labels and then memorize noisy labels. 

        Moreover, we notice a rapid increment of ODS-F and mAP on noisy labels at about 20 epochs when metrics on clean labels are saturated. According to ADELE~\shortcite{liu2022adaptive}, we select the warm-up model with the rapid increment of metrics on the training set. We further check the performance of models at around 20 epochs and find they perform well on clean labels, which validates the rationality of this strategy. For Cityscapes without clean labels, we employ the same setting and select models around 20 epochs as our warm-up model which also performs well on the test set. 

    % Cityscapes performance table
    \begin{table*}
    \centering
        \renewcommand{\arraystretch}{1.0}
        \resizebox{\textwidth}{!}{
        \begin{tabular}{c|c|c|ccccccccccccccccccc|c}
            \toprule
            Setting    & Metric   & Method   & road     & sidewalk     & building     & wall     & fence     & pole     & t-light     & t-sign     & veg     & terrain     & sky     & person     & rider     & car     & truck     & bus     & train     & motor     &  bike     & AVG.     \\
            \midrule
            % CASENet
            \multirow{10}{*}{Thin}    & \multirow{5}{*}{ODS-F}   & CASENet   & 0.862     & 0.748     & 0.748     & 0.482     & 0.468     & 0.729     & 0.703     & 0.735     & 0.794     & 0.567     & 0.865     & 0.806     & 0.666     & 0.883     & 0.495     & 0.672     & 0.498     & \bf 0.566     &  0.718     & 0.684     \\
            % CASENet-S
            ~    & ~   & CASENet-S   & 0.876     & 0.771     & 0.762     & 0.492     & 0.467     & 0.755     & 0.715     & 0.753     & 0.806     & 0.597     & 0.867     & 0.817     & 0.683     & \bf 0.893     & 0.513     & 0.688     & 0.442     & 0.552     &  0.726     & 0.693     \\
            % SEAL
            ~    & ~   & SEAL   & 0.878     & \bf 0.777     & 0.763     & 0.481     & 0.468     & 0.756     & 0.713     & 0.754     & 0.810     & 0.602     & 0.873     & 0.819     & 0.690     & 0.891     & 0.508     & \bf 0.690     & 0.450     & 0.540     &  0.727     & 0.695     \\
            % STEAL
            ~    & ~   & STEAL$\ssymbol{2}$   & \bf 0.879     & \bf 0.777     & \bf 0.773     & \bf 0.494     & \bf 0.491     & \bf 0.790     & \bf 0.746     & \bf 0.764     & \bf 0.821     & \bf 0.597     & \bf 0.881   & 0.808     & 0.701     & 0.836     & \bf 0.519     & 0.675     & \bf 0.531     & 0.557     &  \bf 0.742     & \bf 0.704     \\
            \cline{3-23}
            % Ours
            % ~    & ~   & Ours   & 0.865     & 0.736     & 0.737     & 0.452     & 0.422     & 0.735     & 0.636     & 0.717     & 0.777     & 0.570     & 0.818     & 0.792     & 0.659     & 0.870     & 0.446     & 0.632     & 0.400     & 0.491     &  0.709     & 0.657     \\
            ~    & ~   & Ours   & 0.878     & 0.775     & 0.766     & \bf 0.494     & 0.473     & 0.787     & 0.734     & 0.756     & 0.810     & 0.591     & 0.866     & \bf 0.820     & \bf 0.705     & 0.887     & 0.482     & 0.680     & 0.433     & 0.552     &  0.741     & 0.696     \\
            \cline{2-23}
            % CASENet-MAP
            ~    & \multirow{5}{*}{mAP}   & CASENet   & 0.542     & 0.640     & 0.661     & 0.393     & 0.372     & 0.580     & 0.606     & 0.680     & 0.693     & 0.477     & 0.737     & 0.697     & 0.583     & 0.672     & 0.406     & 0.550     & 0.387     & 0.492     &  0.603     & 0.567     \\
            % CASENet-S-mAP
            ~    & ~   & CASENet-S   & 0.894     & 0.757     & 0.736     & \bf 0.438     & 0.390     & 0.665     & 0.672     & 0.745     & 0.774     & 0.541     & 0.822     & 0.785     & 0.619     & 0.884     & 0.449     & 0.688     & 0.360     & 0.485     &  0.675     & 0.651     \\
            % SEAL-mAP
            ~    & ~   & SEAL   & 0.772     & 0.762     & 0.759     & 0.433     & 0.401     & 0.657     & 0.684     & 0.762     & 0.788     & 0.565     & 0.824     & 0.809     & 0.639     & 0.859     & \bf 0.459     & 0.688     & 0.384     & 0.488     &  0.705     & 0.655     \\
            % STEAL-mAP
            ~    & ~   & STEAL$\ssymbol{2}$   & \bf 0.907     & \bf 0.800     & \bf 0.801     & 0.409     & 0.413     & \bf 0.800     & \bf 0.755     & 0.778     & \bf 0.848     & 0.569     & \bf 0.900     & 0.835     & 0.687     & 0.831     & 0.436     & 0.660     & \bf 0.457     &  0.511     &  0.757     & 0.692     \\
            \cline{3-23}
            % Ours-mAP
            % ~    & ~   & Ours   & 0.880     & 0.750     & 0.724     & 0.380     & 0.336     & 0.666     & 0.608     & 0.725     & 0.759     & 0.515     & 0.787     & 0.784     & 0.612     & 0.880     & 0.371     & 0.632     & 0.304     & 0.439     &  0.681     & 0.623     \\
            ~    & ~   & Ours   & 0.905     & \bf 0.800     & 0.796     & 0.432     & \bf 0.419     & 0.794     & 0.749     & \bf 0.782     & 0.842     & \bf 0.578     & 0.880     & \bf 0.859     & \bf 0.705     & \bf 0.920     & 0.445     & \bf 0.690     & 0.373     & \bf 0.525     &  \bf 0.774     & \bf 0.698     \\
            \midrule % -------------------------------------------------------------------------------------
            % CASENet-raw F
            \multirow{10}{*}{Raw}    & \multirow{5}{*}{ODS-F}   & CASENet   & 0.641     & 0.585     & 0.664     & 0.311     & 0.325     & 0.712     & 0.621     & 0.645     & 0.708     & 0.438     & 0.782     & 0.722     & 0.565     & 0.753     & 0.328     & 0.471     & 0.281     & 0.431     &  0.609     & 0.558     \\
            % CASENet-S-raw-F
            ~    & ~   & CASENet-S   & 0.774     & 0.692     & 0.704     & 0.358     & 0.351     & 0.740     & 0.654     & 0.680     & 0.747     & 0.512     & 0.804     & 0.759     & 0.596     & 0.829     & 0.358     & 0.528     & 0.293     & 0.424     &  0.643     & 0.602     \\
            % SEAL-raw-F
            ~    & ~   & SEAL   & \bf 0.822     & 0.717     & 0.728     & 0.356     & 0.353     & 0.762     & 0.665     & 0.694     & 0.773     & \bf 0.538     & 0.830     & 0.777     & 0.620     & 0.850     & \bf 0.367     & \bf 0.544     & 0.308     & 0.424     &  0.662     & 0.620     \\
            % line
            ~    & ~   & STEAL$\ssymbol{2}$   & 0.758     & 0.685     & 0.699     & 0.349     & 0.361     & 0.734     & 0.668     & 0.677     & 0.735     & 0.497     & 0.787     & 0.729     & 0.591     & 0.765     & 0.353     & 0.528     & \bf 0.377     & 0.438     &  0.638     & 0.598     \\
            \cline{3-23}
            % line
            % ~    & ~   & Ours   & 0.771     & 0.679     & 0.704     & 0.320     & 0.313     & 0.740     & 0.572     & 0.683     & 0.756     & 0.492     & 0.772     & 0.756     & 0.585     & 0.824     & 0.308     & 0.489    & 0.248     & 0.381     &  0.636     & 0.581     \\
            ~    & ~   & Ours   & 0.807     & \bf 0.730     & \bf 0.763     & \bf 0.367     & \bf 0.373     & \bf 0.785     & \bf 0.701     & \bf 0.727     & \bf 0.804     & 0.524     & \bf 0.837     & \bf 0.805     & \bf 0.657     & \bf 0.866     & 0.347     & 0.542    & 0.288     & \bf 0.449     &  \bf 0.714     & \bf 0.636     \\
            \cline{2-23}
            % CASEnet raw mAP
            ~    & \multirow{5}{*}{mAP}   & CASENet   & 0.479     & 0.568     & 0.684     & 0.228     & 0.236     & 0.741     & 0.624     & 0.653     & 0.744     & 0.380     & 0.797     & 0.766     & 0.565     & 0.721     & 0.215     & 0.364     & 0.202     & \bf 0.382     &  0.647     & 0.526     \\
            % CASENet-S-RAW-Map
            ~    & ~   & CASENet-S  & \bf 0.819     & 0.732     & 0.755     & \bf 0.287     & 0.254     & 0.798     & 0.686     & 0.716     & 0.810     & 0.489     & 0.813     & 0.816     & 0.598     & 0.893     & 0.255     & 0.499     & 0.185     & 0.367     &  0.698     & 0.604     \\
            % seal-raw-mAP
            ~    & ~   & SEAL   & 0.780     & 0.740     & 0.777     & 0.277     & 0.256     & 0.796     & 0.686     & 0.732     & 0.834     & \bf 0.508     & 0.825     & 0.828     & 0.621     & 0.888     & \bf 0.257     & \bf 0.507     & 0.195     & 0.356     &  0.717     &  0.609     \\
            % steal-raw mAP
            ~    & ~   & STEAL$\ssymbol{2}$   & 0.811     & 0.728     & 0.750     & 0.263     & 0.263     & 0.791     & 0.693     & 0.719     & 0.799     & 0.472     & 0.804     & 0.788     & 0.581     & 0.811     & 0.244     & 0.490     & \bf 0.278     & 0.377     &  0.683     & 0.597     \\
            \cline{3-23}
            % line
            % ~    & ~   & Ours   & 0.806     & 0.709     & 0.746     & 0.226     & 0.206     & 0.784     & 0.561     & 0.708     & 0.806     & 0.573     & 0.754     & 0.805     & 0.573     & 0.885     & 0.192     & 0.436     & 0.135     & 0.306     &  0.687     & 0.567     \\
            ~    & ~   & Ours   & 0.818     & \bf 0.745     & \bf 0.793     & 0.258     & \bf 0.271     & \bf 0.817     & \bf 0.708     & \bf 0.752     & \bf 0.846     & 0.477     & \bf 0.830     & \bf 0.837     & \bf 0.645     & \bf 0.911     & 0.254     & 0.492     & 0.187     & 0.373     &  \bf 0.755     & \bf 0.619     \\
            \bottomrule
        \end{tabular}
        }
        \caption{Comparisons with previous SOTA methods on Cityscapes. ``AVG.'' and $\ssymbol{2}$ denote category-averaged performance and using NMS for post-processing, respectively.} % $*$ denotes our reproduced results. Although our method obtains relatively low ODS-F and mAP, we emphasize it is reasonable (refer to Sec.\ref{sec:comparison}) under setting of large matching tolerance. Visualized results in Fig.\ref{fig:city_vis} validate our method can generate more precise edge maps than others. 
        \label{tab:cityscapes}
    \end{table*}
    
    \begin{figure*}[t]
        \centering
        \includegraphics[width=6.0 in]{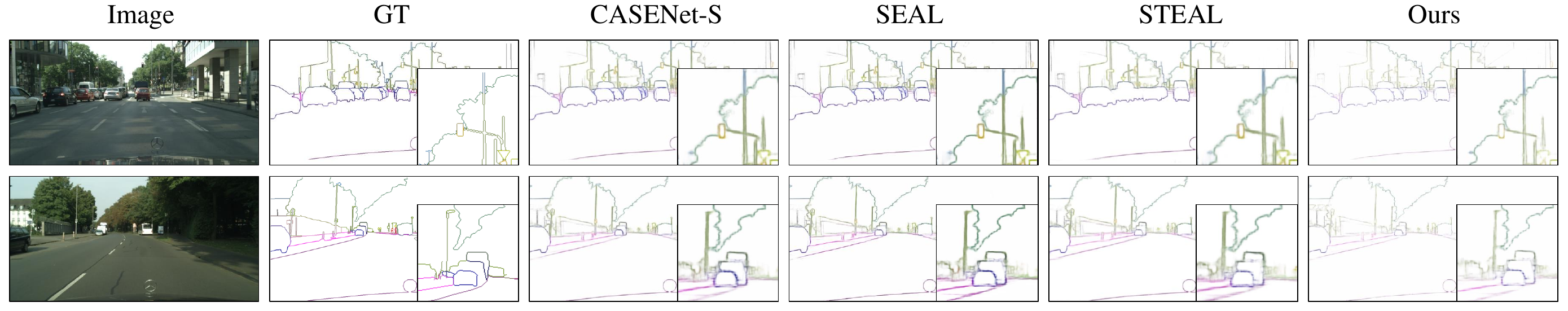}
        \caption{Example results on Cityscapes. Our method obtains much thinner and more accurate edge maps than other methods. }
        \label{fig:city_vis}
    \end{figure*}

    \subsection{Comparison with previous SOTA Methods}
    \label{sec:comparison}    
        %%% original version
        % We compare our method with CASENet, SEAL, and STEAL on SBD and Cityscapes. The baseline is CASENet-S, {\it i.e.,} CASENet with the class-unweighted loss. Note that both SEAL and STEAL are also built on CASENet-S. We emphasize that our PNT-Edge model does not employ NMS for post-processing.  
        %%% updated 0730
        We compare our method with CASENet, SEAL, and STEAL on SBD and Cityscapes. The baseline is CASENet-S, {\it i.e.,} CASENet with the class-unweighted loss. Note that SEAL, STEAL, and our method are all built on CASENet-S with the ResNet-101 backbone. Our PNT-Edge model does not employ NMS for post-processing, as we only observe very slight improvements in our experiments. More discussions about the impact of NMS can be found in our Appendix. 
        %  according to Appendix A.2
    
\begin{figure}[t]
            \centering
            \includegraphics[width=2.8 in]{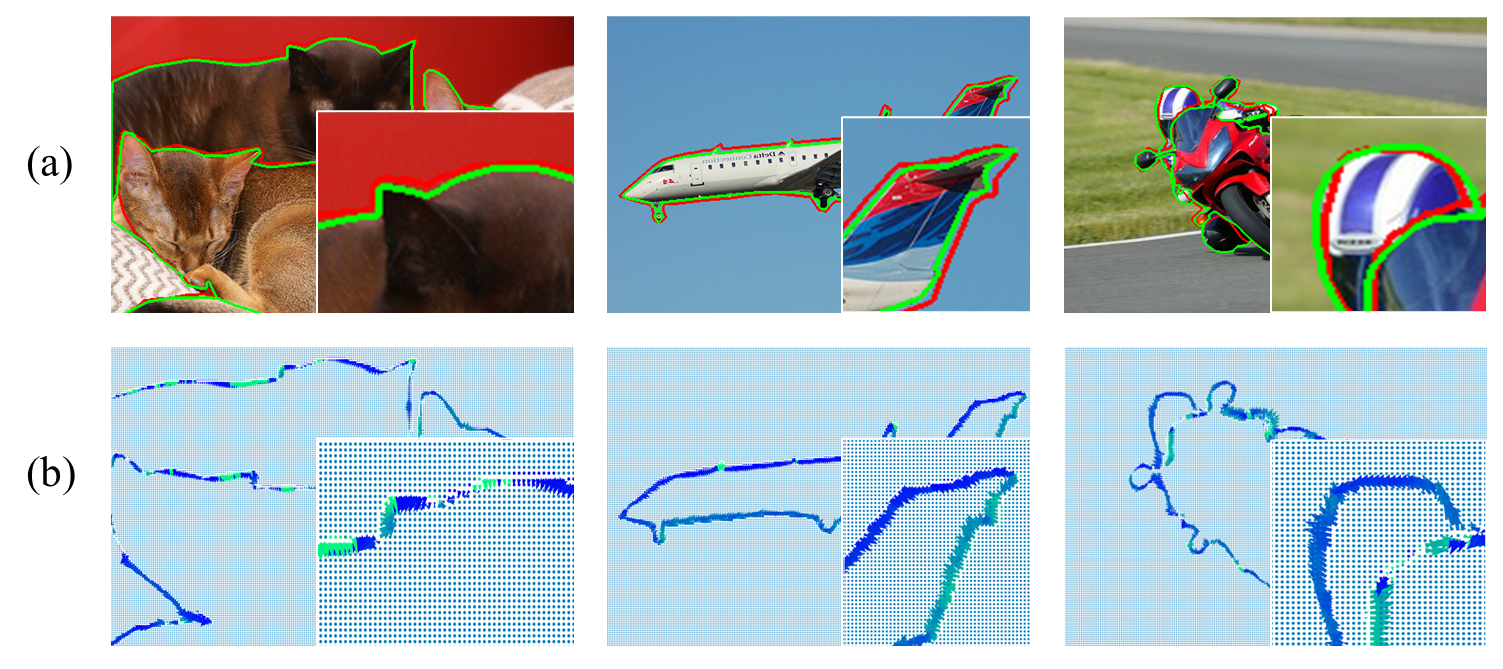}
            \caption{(a) SBD examples with both clean({\color{green}green}) and noisy labels({\color{red}red}). (b) Visualized pixel-level noise transitions. }
            \label{fig:shift_vis}
        \end{figure}

        \paragraph{Performance on SBD} The performance on re-annotated test set is reported in Tab. \ref{tab:sbd}. Our method achieves the highest ODS-F under both settings, surpassing the baseline by 2.4\% and 4.7\%. It also outperforms SEAL ODS-F by 1.3\% and mAP by 1.6\% under ``Thin'' setting. Under ``Raw'' setting, our method obtains 2.3\% higher ODS-F but lower mAP than SEAL (0.653 {\it v.s.} 0.657). Since SEAL generates thicker edges than ours as shown in Fig. \ref{fig:sbd_vis}, it causes better recall but worse precision. Moreover, our method obtains competitive results with STEAL which employs NMS for post-processing. Qualitative results in Fig. \ref{fig:sbd_vis} further validate the effectiveness of our method. Fig. \ref{fig:shift_vis} visualizes predicted noise transitions, {\it i.e.,} displacement fields. 

        % v1
        % \paragraph{Performance on Cityscapes} Following SEAL, we set the matching tolerance as 0.0035 ($\approx$ 8 pixels), and test with original labels that contain low-level label noise. The results are listed in Table \ref{tab:cityscapes}. Though our method obtains relatively a little lower ODS-F and mAP, we emphasize it does not mean our method fails. This is because, with large tolerance, blurred edges would be considered as true positives as well, which brings high precision and recall. As visualized in Fig.\ref{fig:city_vis}, since other methods including the baseline predict more blurred edge maps than ours, it is reasonable to achieve higher performance. Moreover, the predictions of our method contain more precise edges and fit ground truth better because of relieving the impact of label noise. Thus, it is more rational to employ stricter evaluation metrics, {\it i.e.} small matching tolerance, for evaluating our thinner edge maps of ours.

        \paragraph{Performance on Cityscapes} Following SEAL, we test on original labels that contain low-level label noise. The results are listed in Tab. \ref{tab:cityscapes}. Under ``Thin'' setting, our method obtains competitive performance on ODS-F compared with the baseline and previous SOTA approaches because of little label noise on the Cityscapes. But we notice that PNT-Edge surpasses the baseline mAP enormously by 4.7\%. This is because our method can fit ground truth better with few blurs as visualized in Fig. \ref{fig:city_vis}. Moreover, our method achieves the best performance on both ODS-F and mAP under ``Raw'' setting, which proves PNT-Edge can produce precise edge maps even without post-processing. Our method performs best under even stricter criteria as delivered in Sec. \ref{sec: discussions}.
        % set the matching tolerance as 0.0035 ($\approx$8 pixels), and 
    
    %------------------------------------------------------------------------    
    \subsection{Ablation Study}
    \label{sec:ablation}
        We validate the impact of hyper-parameters and training strategies on SBD. We conduct experiments on the original validation set (Val-noisy) with noisy labels and a 50\% random sampled re-annotated test set (Val-clean) with clean labels. The baseline is CASENet-S.

        \paragraph{Impact of confident threshold $\tau$ and loss weight $\alpha_4$ for $L_{dns}$} To analyze the effect of confident-sample extraction and local edge density regularization, we experiment with different settings of $\tau$ and $\alpha_4$ and report in Tab. \ref{tab:ablation-tau}. For one thing, a lower $\tau$ leads to higher ODS-F on both noisy and clean labels, but a too-low $\tau$ degrades mAP remarkably, especially on clean labels. The reasons are two folds. First, the confidence of predicted edge maps is relatively low because of unweighted loss functions. Second, more confident examples help learn better noise transitions, but too low $\tau$ would introduce noisy examples and degrade performance. For another, when $\alpha_4$ increases, ODS-F increases consistently. However, mAP improves first and then decreases on Val-noisy and Val-clean. To make a trade-off between ODS-F and mAP on the clean labels, we choose $\tau=0.1, \alpha_4=3$ for experiments, where our method outperforms the baseline ODS-F by 1.3\% and mAP by 4.4\%. This fact verifies the effectiveness of our method. Note that though better performance on noisy labels indicates probably better performance on clean labels, it is not always the fact. It is possible to take metrics on noisy labels for reference only when clean labels are unavailable and the label noise is at a low level.

        \begin{table}[t]
            \centering
            \scalebox{0.7}{
                \begin{tabular}{c|c|cc|cc}
                    \toprule
                    \multirow{2}{*}{$\tau$}   & \multirow{2}{*}{$\alpha_4$}     & \multicolumn{2}{c}{Val-noisy} \vline  & \multicolumn{2}{c}{Val-clean}\\
                     ~       & ~    & ODS-F    & mAP   & ODS-F    & mAP\\
                    \hline
                    \multicolumn{2}{c}{Baseline} \vline & 0.722   & 0.710    & 0.653   & 0.579\\
                    \hline
                    \multirow{5}{*}{$0.01$}    & $0.0$    & \bf 0.736   & 0.719   & \bf 0.672   &  0.590\\
                    ~    & $1.0$    & 0.729   & \bf 0.721   & 0.669   & \bf 0.613\\
                    ~    & $2.0$    & 0.729   & \bf 0.721   & 0.669   & 0.612\\
                    ~    & $3.0$    & 0.729   & 0.720   & 0.666   & 0.607\\
                    ~    & $5.0$    & 0.729   & 0.717   & 0.659   & 0.586\\

                    \hline
                    \multirow{5}{*}{\textcolor{red}{$0.1$}}    & $0.0$    & 0.717   & 0.706   & 0.652   &  0.605\\
                    ~    & $1.0$    & 0.718   & 0.709   & 0.651   & 0.607\\
                    ~    & $2.0$    & 0.723   & 0.717   & 0.662   & \bf 0.626\\
                    ~    & \textcolor{red}{$3.0$}    & \textcolor{red}{\bf 0.727}   & \textcolor{red}{\bf 0.723}   & \textcolor{red}{0.666}   & \textcolor{red}{0.623}\\
                    ~    & $5.0$    & \bf 0.727   & 0.721   & \bf 0.669   & 0.621\\
                    \hline
                    $0.2$    & $0.0$    & 0.691   & 0.677   & 0.617   &  0.560\\  
                    \bottomrule
                \end{tabular}
            }
            \caption{Discussions of confident threshold $\tau$ and loss weight $\alpha_4$ for $L_{dns}$. \textcolor{red}{Red} indicates the trade-off setting.}
            \label{tab:ablation-tau}
        \end{table}

        \begin{table}[t]
            \centering
            \scalebox{0.8}{
            \begin{tabular}{c|cc|cc}
                \toprule
                \multirow{2}{*}{Strategy}      & \multicolumn{2}{c}{Val-noisy} \vline  & \multicolumn{2}{c}{Val-clean}\\
                ~          & ODS-F    & mAP   & ODS-F    & mAP\\
                \hline
                Label Correction  & 0.678   & 0.653   & 0.610    & 0.537 \\
                Joint Training    & \bf 0.717   & \bf 0.706   & \bf 0.652    & \bf 0.605\\
                \bottomrule
            \end{tabular}
            }
            \caption{Comparison of different training strategies on SBD. }
            \label{tab:ablation-strategy}
        \end{table}

        \paragraph{Joint Training {\it v.s.} Label Correction} To further validate the effectiveness of our method, we compare two kinds of training strategies: joint training and label correction. The joint training is delineated in Sec. \ref{sec:tedge}. For label correction, we employ $\phi^{-1}$ to refine noisy labels and then train CASENet-S with refined labels. Referring to Tab. \ref{tab:ablation-strategy}, joint training is proved to be more effective than label correction. Because PSL module would dynamically adapt noisy labels through joint training. Besides, generating refined labels for the whole dataset is time-consuming. As a result, we employ the joint training strategy in PNT-Edge. 

        \paragraph{Effect of loss functions} As Tab. \ref{tab:ablation-reg} shows, all loss functions help improve the performance on Val-clean and Val-noisy, but smooth loss $L_{smth}$ shows useless. Because the local edge density regularization, which encourages similar displacements on pixels with similar local edge densities, also works for smooth regularization. Tab. \ref{tab:ablation-reg} also verifies the effectiveness of the proposed local edge density regularization term in pixel-level noise-transition identification. Moreover, the unmatched loss plays an important role in producing precise edge maps with high mAP.

        \begin{table}[t]
            \centering
            \scalebox{0.7}{
                \begin{tabular}{cccc|cc|cc|cc}
                    \toprule
                    \multicolumn{4}{c}{PSL} \vline &  \multicolumn{2}{c}{Joint} \vline & \multicolumn{2}{c}{Val-noisy} \vline  & \multicolumn{2}{c}{Val-clean}\\
                    $L_{sup}$  &  $L_{sim}$  &  $L_{smth}$  &  $L_{dns}$  &  $L_{edge}$  &  $L_{um}$  & ODS-F    & mAP   & ODS-F    & mAP\\
                    \hline
                    $\checkmark$  &  ~  &  ~  &  ~  &  $\checkmark$  &  $\checkmark$  & 0.679   & 0.659  & 0.603  & 0.538\\
                    $\checkmark$  &  ~  &  $\checkmark$  &  $\checkmark$  &  $\checkmark$  &  $\checkmark$  & 0.716   & 0.703  & 0.642  & 0.586\\
                    \textcolor{red}{$\checkmark$}  &  \textcolor{red}{$\checkmark$}  &  ~  &  \textcolor{red}{$\checkmark$}  &  \textcolor{red}{$\checkmark$} &  \textcolor{red}{$\checkmark$}  & \textcolor{red}{0.726}   & \textcolor{red}{0.720}  & \textcolor{red}{\bf 0.668}  & \textcolor{red}{\bf 0.627}\\
                    $\checkmark$  &  $\checkmark$  &  $\checkmark$  &  ~  &  $\checkmark$  &  $\checkmark$  & 0.717   & 0.706  & 0.652  & 0.605\\
                    \hline
                    $\checkmark$  &  $\checkmark$  &  $\checkmark$  &  $\checkmark$  &  $\checkmark$  &  ~  & 0.723   & 0.461  & 0.659  & 0.369\\
                    \hline
                    $\checkmark$  &  $\checkmark$  &  $\checkmark$  &  $\checkmark$  &  $\checkmark$  &  $\checkmark$  & \bf 0.727   & \bf 0.723  & 0.666  & 0.623\\
                    \bottomrule
                \end{tabular}
            }
            \caption{Effects of loss functions. We are more concerned about the results on Val-clean since our goal is to predict clean edges. \textcolor{red}{Red} indicates our experiment setting. }
            \label{tab:ablation-reg}
        \end{table}

        \subsection{Discussions}
        \label{sec: discussions}

            \begin{figure}[t]
                \centering
                \includegraphics[width=2.5 in]{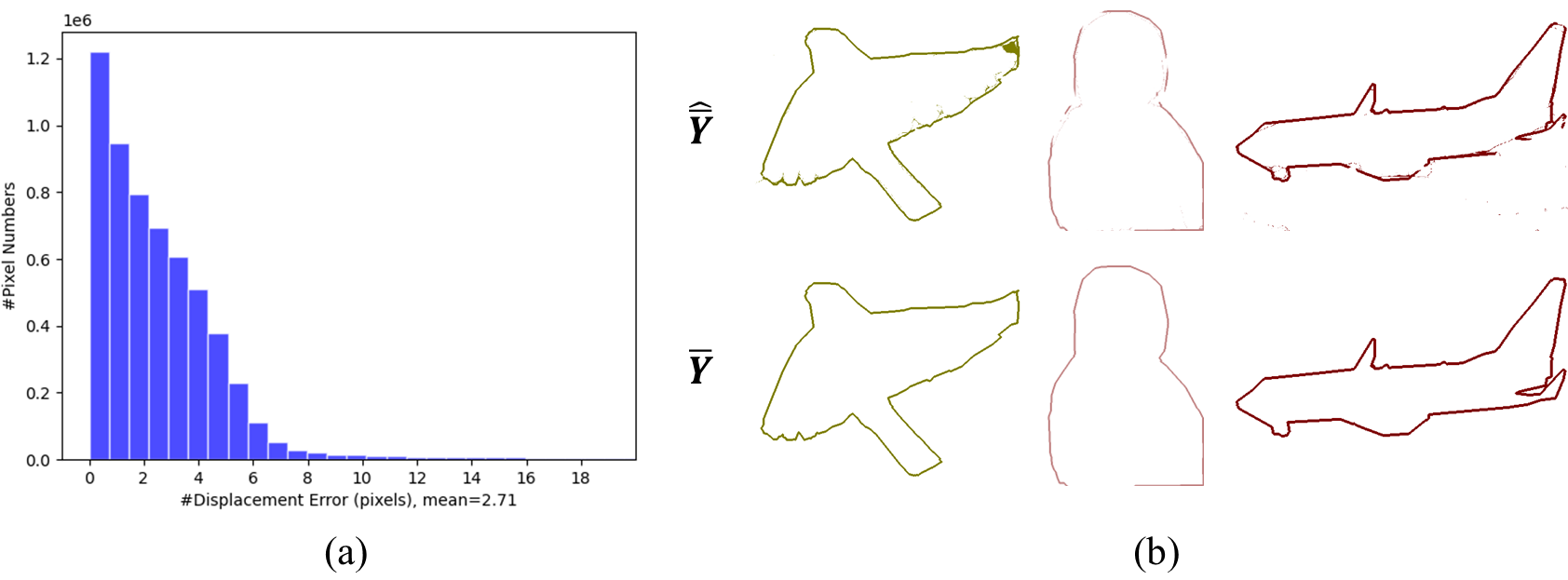}
                \caption{Error analysis. (a) Error histogram between predicted transitions with min-distance-matching results. (b) Comparison of transformed prediction $\hat{\bar{Y}}$ and noisy label $\bar{Y}$.}   % Error analysis for predicted noise transitions.
                \label{fig:check}
            \end{figure}
            
            % \paragraph{Noise-transition error analysis} Since ground truth noise transitions are not available, we analyze the error of predicted noise transitions {\it i.e.} displacement field $\phi$ from two aspects. First, we refer to the min-distance-matching results. In Fig.\ref{fig:check}(a), the error histogram reports an average error of 2.71 pixels. It is reasonable because of the gaps between min-distance-matching results and ideal ground-truth transitions. Second, considering that ideal transitions will produce high-quality predictions of the noisy label $\bar{Y}$, we compare the transformed edge prediction $\hat{\bar{Y}}$ with noisy label $\bar{Y}$. The ODS-F of $\hat{\bar{Y}}$ achieve 0.790 on SBD under ``RAW'' setting, indicating the predicted transition $\phi$ model label corruption well as shown in Fig.\ref{fig:check}(b). Visualized displacement fields refer to Fig.\ref{fig:shift_vis}.
            \paragraph{Noise-transition error analysis} To analyze the error of the predicted noise transition, {\it i.e.,} displacement field $\phi$, we first compare with the min-distance-matching result on SBD. The error histogram in Fig. \ref{fig:check}(a) reports an average error of 2.71 pixels. It is reasonable because of the gaps between min-distance-matching results and the ideal ground-truth transitions. Furthermore, considering that ideal transitions will produce high-quality predictions of the noisy label $\bar{Y}$, we compare the transformed edge prediction $\hat{\bar{Y}}$ with noisy label $\bar{Y}$. The ODS-F of $\hat{\bar{Y}}$ achieve 0.790 under ``RAW'' setting, indicating the predicted transition $\phi$ models label corruption well as shown in Fig. \ref{fig:check}(b). Visualized displacement fields refer to Fig. \ref{fig:shift_vis}.       

            % \paragraph{Model size and computational complexity} Since PSL is only an auxiliary module for tackling label noise via modeling label corruption during the training process, our method does not bring additional computation for inference. Tab. \ref{tab: complexity} compares the model size and computational complexity with the baseline CASENet-S for model training. Our PNT-Edge provides an effective training strategy to tackle noisy edge labels at the cost of negligible 0.16M more parameters and 15.4\% FLOPs additional computation.

%--------------------------------------------------------------------------------------------------------------------------------------------
\section{Conclusion}
\label{sec:conclusion}
    % This paper proposes {\bf PNT-Edge} to tackle label noise for edge detection via modeling label corruption. To achieve it, we develop a PSL module to learn pixel-level noise transitions. And a local edge density regularization is proposed to help estimate such transitions via prior knowledge. Our method proves to be effective, outperforming SEAL ODS-F by 1.3\% and mAP by 1.6\% on SBD with more precise edge maps. Our method also obtains competitive ODS-F and much higher mAP on Cityscapes with low-level label noise. 
    This paper proposes {\bf PNT-Edge} to tackle label noise for edge detection via modeling label corruption. To achieve it, we develop a PSL module to learn pixel-level noise transitions. And a local edge density regularization is proposed to help estimate such transitions via prior knowledge. Our method outperforms SEAL ODS-F by 1.3\% and mAP by 1.6\% on SBD and obtains competitive ODS-F and 4.3\% higher mAP than SEAL on Cityscapes. The proposed PNT-Edge proves to be effective in relieving the impact of label noise and producing more precise edge maps.

%%
%% The acknowledgments section is defined using the "acks" environment
%% (and NOT an unnumbered section). This ensures the proper
%% identification of the section in the article metadata, and the
%% consistent spelling of the heading.
\begin{acks}
    This work was supported in part by the National Natural Science Foundation of China under Grant 62076186 and 62225113. The numerical calculations in this paper have been done on the supercomputing system in Supercomputing Center of Wuhan University.
\end{acks}

%%
%% The next two lines define the bibliography style to be used, and
%% the bibliography file.
\bibliographystyle{ACM-Reference-Format}
\balance
\bibliography{sample-base}
%%
%% If your work has an appendix, this is the place to put it.
\appendix
\section{Appendix}
    \subsection{The Overall Pipeline of PNT-Edge}
    We summarize the overall training pipeline of our PNT-Edge in Alg.\ref{alg:algorithm}. It consists of three steps: a) warm-up training of the edge detector with noisy labels, b) training the PSL module with reliable examples to model the label corruption of noisy labels, and c) joint training of the edge detector and PSL with noisy labels.

    \subsection{More Discussions}
    \paragraph{Discussion of NMS Post-processing} To validate that PNT-Edge can generate crisp edge predictions without NMS, we further conduct experiments with NMS and report the results in Tab.\ref{tab: nms}. While STEAL employed NMS, our method outperforms STEAL by 0.5\% ODS-F and 0.5\% mAP on SBD under the ‘Thin’ setting without NMS. Moreover, we find NMS brings limited increments to our method, which indicates our method directly generates crisp edge predictions without NMS post-processing. On the other hand, since NMS helps to relieve the blurred edge predictions caused by label noise, the results further validate the effectiveness of our method in relieving the impact of label noise.

    \begin{table}[t]
        \centering
        \scalebox{0.9}{
        \begin{tabular}{ccc}
            \toprule
            \textbf{Method} & \textbf{ODS-F} & \textbf{mAP}  \\
            \midrule
            STEAL (NMS)     & 0.677          & 0.622         \\
            % \midrule
            Ours (w/o NMS)  & 0.682          & 0.627         \\
            Ours (NMS)      & 0.685          & 0.627       \\
            \bottomrule
        \end{tabular}}
        \caption{Model size and computational complexity for training. PNT-Edge does not bring additional cost for inference. }
        \label{tab: nms}
    \end{table}

    \paragraph{Performance under stricter criteria} To illustrate advances of our method, Tab. \ref{tab: metric} compares the performance under stricter criteria {\it i.e.,} lower matching threshold as 0.00175($\approx$ 4 pixels) and 0.000875($\approx$ 2 pixels) with previous approaches on Cityscapes. Our PNT-Edge achieves SOTA on both settings, surpassing the baseline ODS-F by 6.9\% and mAP by 7.5\% impressively under a 0.00175 threshold. Under an even stricter threshold of 0.000875, our method also obtains the best performance. These results prove our method produces crisp and precise edge maps effectively. 

    \paragraph{Model size and computational complexity} Since PSL is only an auxiliary module for tackling label noise via modeling label corruption during the training process, our method does not bring additional computation for inference. Tab. \ref{tab: complexity} compares the model size and computational complexity with the baseline CASENet-S for model training. Our PNT-Edge provides an effective training strategy to tackle noisy edge labels at the cost of negligible 0.16M more parameters and 15.4\% FLOPs additional computation.

    \makeatletter
    \makeatother
    \begin{table}[t]
        \centering
        \scalebox{0.8}{
        \begin{tabular}{c|c|cc}
            \toprule
            Threshold  &  Method  & ODS-F(\%) & mAP(\%) \\
            \midrule
            \multirow{5}{*}{0.00175}  &   CASENet     & 0.420          & 0.343        \\
            ~  &  CASENet-S          & 0.457          &   0.436        \\
            ~  &  SEAL          & 0.487          &   0.466        \\
            ~  &  STEAL$\ssymbol{2}$          & 0.444          &   0.413        \\
            \cline{2-4} 
            ~  &  Ours          & \bf 0.526          &  \bf 0.521        \\
            \midrule
            \multirow{5}{*}{0.000875}  &   CASENet     & 0.372          & 0.285        \\
            ~  &  CASENet-S          & 0.404          &   0.359        \\
            ~  &  SEAL          & 0.427          &   0.380        \\
            ~  &  STEAL$\ssymbol{2}$          & 0.396          &   0.347        \\
            \cline{2-4} 
            ~  &  Ours          & \bf 0.453          & \bf 0.418        \\
            \bottomrule
        \end{tabular}}
        \caption{Performance comparison under stricter criteria on Cityscapes. Our method achieves the best performance under stricter criteria, {\it i.e.,} lower matching threshold. } 
        % generates more precise and crisp edge maps and 
        \label{tab: metric}
    \end{table}

    \begin{table}[t]  % h
        \centering
        \scalebox{0.9}{
        \begin{tabular}{ccc}
            \toprule
            Method  & Params(M) & FLOPs(G) \\
            \midrule
            CASENet-S     & 42.44          & 175.6        \\
            Ours          & 42.60          & 202.6        \\
            \bottomrule
        \end{tabular}}
        \caption{Model size and computational complexity for training. PNT-Edge does not bring additional cost for inference. }
        \label{tab: complexity}
    \end{table}

    % Algorithm
    \begin{algorithm}[t]
        \caption{{\it PNT-Edge} for noisy edge labels learning}
        \label{alg:algorithm}
        {\bf Input:} Noisy training set $\bar{D}=\{X_i, \bar{Y_i}\}_{i=1}^N$, threshold $\tau$. \ \ \ \ \ \ \ \ \ \ \ \ \ \ \ \ \ \ \ \ \ \\
        {\bf Output:} Noise-robust edge detector $f(X; \hat{\theta})$. \ \ \ \ \ \ \ \ \ \ \ \ \ \ \ \ \ \ \ \ \ \ \ \ \ \ \ \ \ \ \ \ \ \ \ \ \ \ \ \ 
        
        \begin{algorithmic}[1] %[1] enables line numbers
            \STATE $//$ {\it Step 1: Warm-up training}
            \STATE Train the edge detector on the noisy dataset $\bar{D}$ to obtain the initial detector $f(X; \theta_0)$ with Eq.8.
            \STATE $//$ {\it Step 2: Train the PSL module}
            \FOR{$i=1$ to $N$}
                \STATE Extract confident labels $Y^S_i$ as Eq.5 with $f(X; \theta_0)$; 
                \STATE Generate field $\phi$ and transformed output $\hat{\bar{Y}}$ as Eq.6; 
                \STATE Compute the loss as Eq.12;
                \STATE Optimize the parameter $\psi$ by SGD; 
            \ENDFOR
            \STATE \textbf{return} $\hat{\psi}$
            
            \STATE $//$ {\it Step 3: Train the edge detector with PSL}
            \FOR{$n=1$ to $N$}
                \STATE Generate predictions $\hat{Y}_i, \hat{\bar{Y}}_i$ by $f\circ g$ as Eq.4 and Eq.6; 
                \STATE Compute the loss as Eq.15;
                \STATE Optimize the parameter $\theta$ by SGD; 
            \ENDFOR
            
            \STATE \textbf{return} $\hat{\theta}$
        \end{algorithmic}
    \end{algorithm}

\section{General Edge Detection}
    \subsection{Apply PNT-Edge to General Edge Detection}
    In our paper, we mainly focus on semantic edges for a fair comparison with previous works on noisy edges, i.e., SEAL and STEAL. Since our method is edge-detector-agnostic, it is also easy to apply our method to general edge detection. Here is an instruction.

    \begin{enumerate}
        \item Replace the semantic edge detector (i.e., CASENet) with the general edge detector (i.e., HED and RCF). 
        \item Use the general edge detection loss function, i.e., binary cross-entropy loss, instead of the multi-classification loss. And no more changes are needed.
        \item Follow the same training procedure as Alg.1, and implement warm-up training, PSL-module training, and joint-training of the edge detector with PSL step by step.
    \end{enumerate}

        \begin{table*}[t]
        \centering
        \scalebox{0.9}{
        \begin{tabular}{c|ccc|ccc}
            \toprule
            % \multicolumn{2}{c}{Joint} \vline
            \multirow{2}{*}{Method}  & \multicolumn{3}{c}{Thin} \vline & \multicolumn{3}{c}{Raw} \\
            \cline{2-7}
            ~  & ODS-F  & OIS-F & mAP & ODS-F  & OIS-F & mAP \\
            \midrule
            HED & 0.756  & 0.774  & 0.665  & 0.600  & 0.619  & 0.601 \\
            HED-PNT  & \bf 0.767 \textcolor{red}{\small{$\uparrow 0.011$ }}  & \bf 0.786 \textcolor{red}{\small{$\uparrow 0.012$ }}  & \bf 0.734 \textcolor{red}{\small{$\uparrow 0.069$ }} & \bf 0.654 \textcolor{red}{\small{$\uparrow 0.054$ }}  & \bf 0.665 \textcolor{red}{\small{$\uparrow 0.046$ }}  & \bf 0.698 \textcolor{red}{\small{$\uparrow 0.097$ }} \\
            \midrule
            RCF  & 0.765  & 0.781  & 0.676  & 0.627  & 0.642  & 0.637 \\
            RCF-PNT  & \bf 0.771 \textcolor{red}{\small{$\uparrow 0.006$ }}  & \bf 0.787 \textcolor{red}{\small{$\uparrow 0.007$ }}  & \bf 0.723 \textcolor{red}{\small{$\uparrow 0.047$ }} & \bf 0.642 \textcolor{red}{\small{$\uparrow 0.015$ }} & \bf 0.651 \textcolor{red}{\small{$\uparrow 0.009$ }} & \bf 0.684 \textcolor{red}{\small{$\uparrow 0.047$ }}\\
            \bottomrule
            \end{tabular}}
        \caption{Performance comparison of our method, {\it i.e., } HED-PNT and RCF-PNT, with their baseline models on BSDS500. Our method brings consistent improvements to the baseline under all settings for the general-edge detection task.} 
        \label{tab: bsds}
    \end{table*}

    \begin{figure*}[t]
        \centering
        \includegraphics[width=5.5 in]{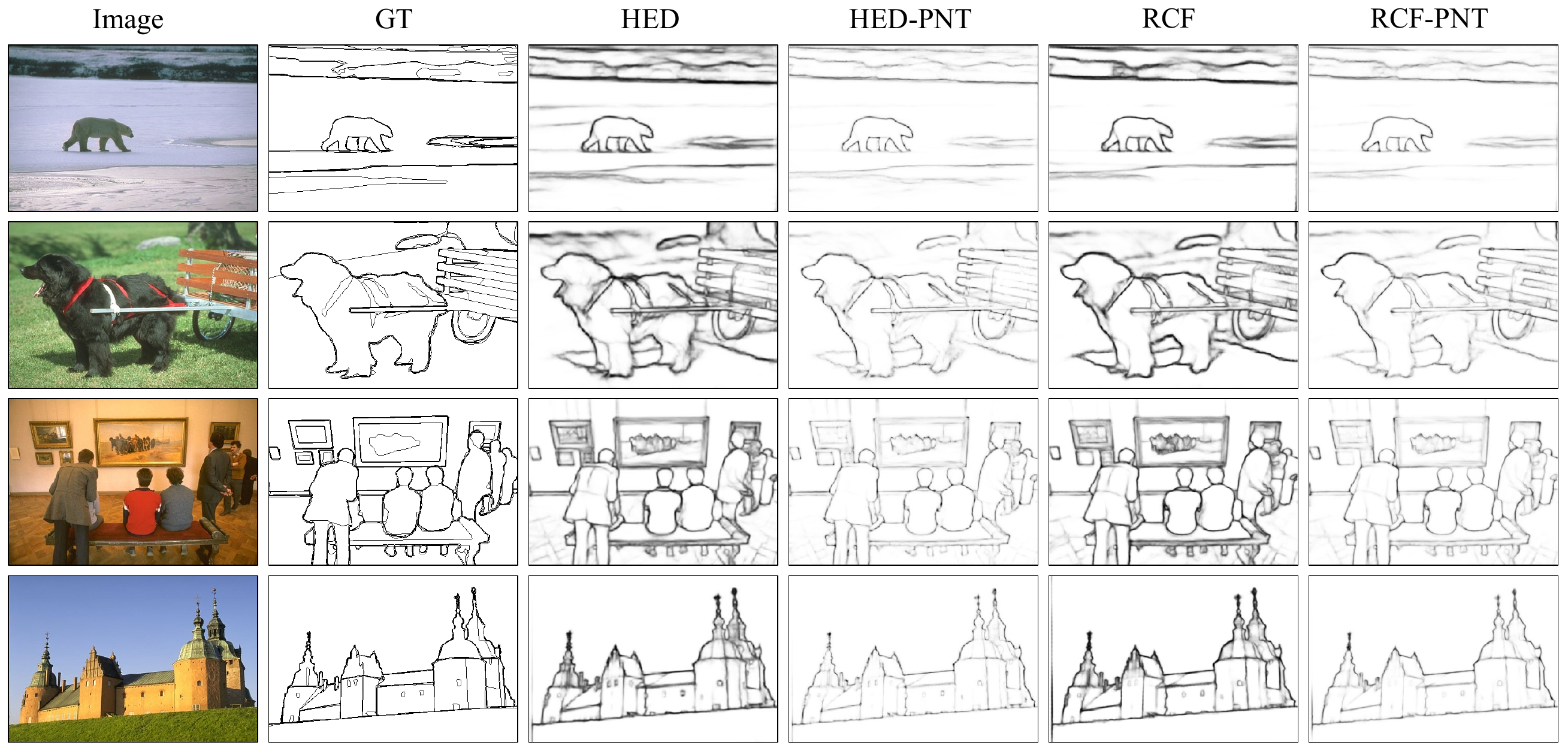}
        \caption{Example results of our PNT-Edge model and the baseline on BSDS500. Our method can produce more precise edge maps for general edge detection by relieving the label noise issue. }
        \label{fig:bsds_vis}
    \end{figure*}

    \subsection{Results on BSDS500}
    We conduct experiments on BSDS500 to explore the effectiveness of our PNT-Edge in general edge detection for label noise. {\bf Since BSDS500 does not provide clean labels for test, we emphasize that the experimental results here, which are calculated under a matching tolerance with the ground truth, just give a reference for the performance of our method in detecting general edges with label noise.} 

    \paragraph{BSDS500 Dataset} BSDS500, which is released by Berkley in 2013, is the most widely used benchmark for general edge detection. It contains 200 train images, 100 validate images and 200 test images. Each image contains 5 to 7 annotations by different annotators. So, it contains misaligned edge labels and annotator bias. Note that BSDS500 does not provide clean labels. The train-val set is used for training. And the data augmentation is the same as HED. 

    \newpage
    \paragraph{Experimental Settings} To verify the effectiveness of relieving the label-noise impact, we choose two representative methods, HED and RCF, as our general-edge detectors on BSDS500. We first train HED and RCF with class-unweighted loss following discussions in SEAL as baseline. Note that we employ labels from all annotators for training. For comparison, we follow the instructions in B.1 and employ the proposed PNT-Edge method to learn the label corruption and finetune the baseline model, {\it i.e., } HED and RCF, to produce label-noise-robust general-edge detectors, denoted as HED-PNT and RCF-PNT. We randomly crop the image by $512\times 512$ for training. For evaluation, we set the matching tolerance to 0.0075 following HED and report the performance under ``Thin'' and ``Raw'' settings without NMS.
    %Note that we follow their original training setting and employ labels from all annotators for training. 
    %, where hyper-parameters are empirically set as $\tau=0.2, \alpha_1=0.01, \alpha_2=1, \alpha_3=0, \alpha_4=3$
    
    \paragraph{Performance on BSDS500} As reported in Tab. \ref{tab: bsds}, our method brings consistent improvements to both HED and RCF under all evaluation settings, especially on mAP. Take HED for example. Our PNT-Edge improves the ODS-F, OIS-F and mAP by 1.1\%, 1.2\% and 6.9\% respectively under ``Thin'' setting. It also brings 5.4\% ODS-F, 4.6\%  OIS-F and 9.7\% mAP increments under ``Raw'' setting. These results further prove that our method can generate more precise edge maps without NMS by relieving the impact of noisy labels. The visualization results shown in Fig. \ref{fig:bsds_vis} also verify the effectiveness of our method in dealing with label noise for general edge detection.

\end{document}